%% file: root.tex
\documentclass[letterpaper, 10 pt, conference]{ieeeconf}  
\usepackage[dvipsnames,table]{xcolor}
\usepackage{adjustbox}
\usepackage{colortbl}
\usepackage{array}
\usepackage{url}
\usepackage{multicol,multirow}
\usepackage{booktabs}
\usepackage{subcaption}
\usepackage{float}
\usepackage{amsmath}
\usepackage{xspace}
\usepackage{graphicx}
\usepackage{tabularx}
\usepackage{multirow}
\usepackage{rotating}
\usepackage{array} 
\let\labelindent\relax
\usepackage{enumitem}
\definecolor{cvprblue}{rgb}{0.21,0.49,0.74}
 \usepackage[pagebackref,breaklinks,colorlinks,allcolors=cvprblue]{hyperref}
\usepackage{pifont}
\usepackage{amssymb}
\usepackage{tikz}

\usepackage{booktabs,multirow,array}

\newcolumntype{Y}{>{\centering\arraybackslash\bfseries}X}

\newcommand{\Larrow}{\mathrel{\tikz[baseline=-0.5ex]{\draw[->] (0,0.7ex) -- (0,0) -- (1.2em,0);}}}

\IEEEoverridecommandlockouts                              

\newcolumntype{H}{>{\setbox0=\hbox\bgroup}c<{\egroup}@{}}

\title{\LARGE \bf
BLAZER: \underline{B}ootstrapping \underline{L}LM-based Manipulation \underline{A}gents \\with \underline{Zer}o-Shot Data Generation}

\author{Rocktim Jyoti Das$^{*1}$, Harsh Singh$^{*1}$, Diana Turmakhan$^{\dagger1}$,  Muhammad Abdullah Sohail$^{\dagger1}$, Mingfei Han$^{1}$, \\ Preslav Nakov$^{1}$, Fabio Pizzati$^{1}$, Ivan Laptev$^{1}$%
\thanks{*Co-first author.}%
\thanks{$\dagger$Co-second author.}%
\thanks{$^{1}$Mohamed bin Zayed University of Artificial Intelligence, UAE}%
\thanks{email: firstname.lastname@mbzuai.ac.ae,}%
\thanks{Project Page: \protect\url{https://blazer-llm-agent.github.io/}}%
}

\newcommand{\framework}{{{BLAZER}}\xspace}

\begin{document}

\maketitle
\thispagestyle{empty}
\pagestyle{empty}

\begin{abstract}
Scaling data and models has played a pivotal role in the remarkable progress of computer vision and language. Inspired by these domains, recent efforts in robotics have similarly focused on scaling both data and model size to develop more generalizable and robust policies. 
However, unlike vision and language, robotics lacks access to internet-scale demonstrations across diverse robotic tasks and environments. 
As a result, the scale of existing datasets typically suffers from the need for manual data collection and curation.
To address this problem, here we propose \framework, a framework that learns manipulation policies from \textit{automatically generated training data}. 
We build on the zero-shot capabilities of LLM planners and automatically generate demonstrations for diverse manipulation tasks in simulation. Successful examples are then used to finetune an LLM and to improve its planning capabilities without human supervision. 
Notably, while BLAZER training requires access to the simulator's state, we demonstrate direct transfer of acquired skills to sensor-based manipulation.
Through extensive experiments, we show BLAZER to significantly improve zero-shot manipulation in both simulated and real environments. 
Moreover, BLAZER improves on tasks outside of its training pool and enables downscaling of LLM models. 
Our code and data will be made publicly available on the project page~\cite{blazerwebpage}. 

\end{abstract}

\input{sections/intro}
\input{sections/related}

\input{sections/method}
\input{sections/experiments}
\input{sections/conclusions}

\bibliographystyle{IEEEtran} %
\bibliography{refs}
\newpage
\onecolumn
\appendices
\section*{\textbf{Appendix}}
This appendix provides additional details about our \framework framework, including LLM Agent Prompt in Appendix\ref{app:prompts}, and RLBench tasks in Appendix\ref{app:rlbench_tasks}.
\subsection{Prompts}
\label{app:prompts}
\begin{figure*}[ht!]
  \centering
  \includegraphics[trim={0cm 9cm 0cm 0},clip, width=1\linewidth]{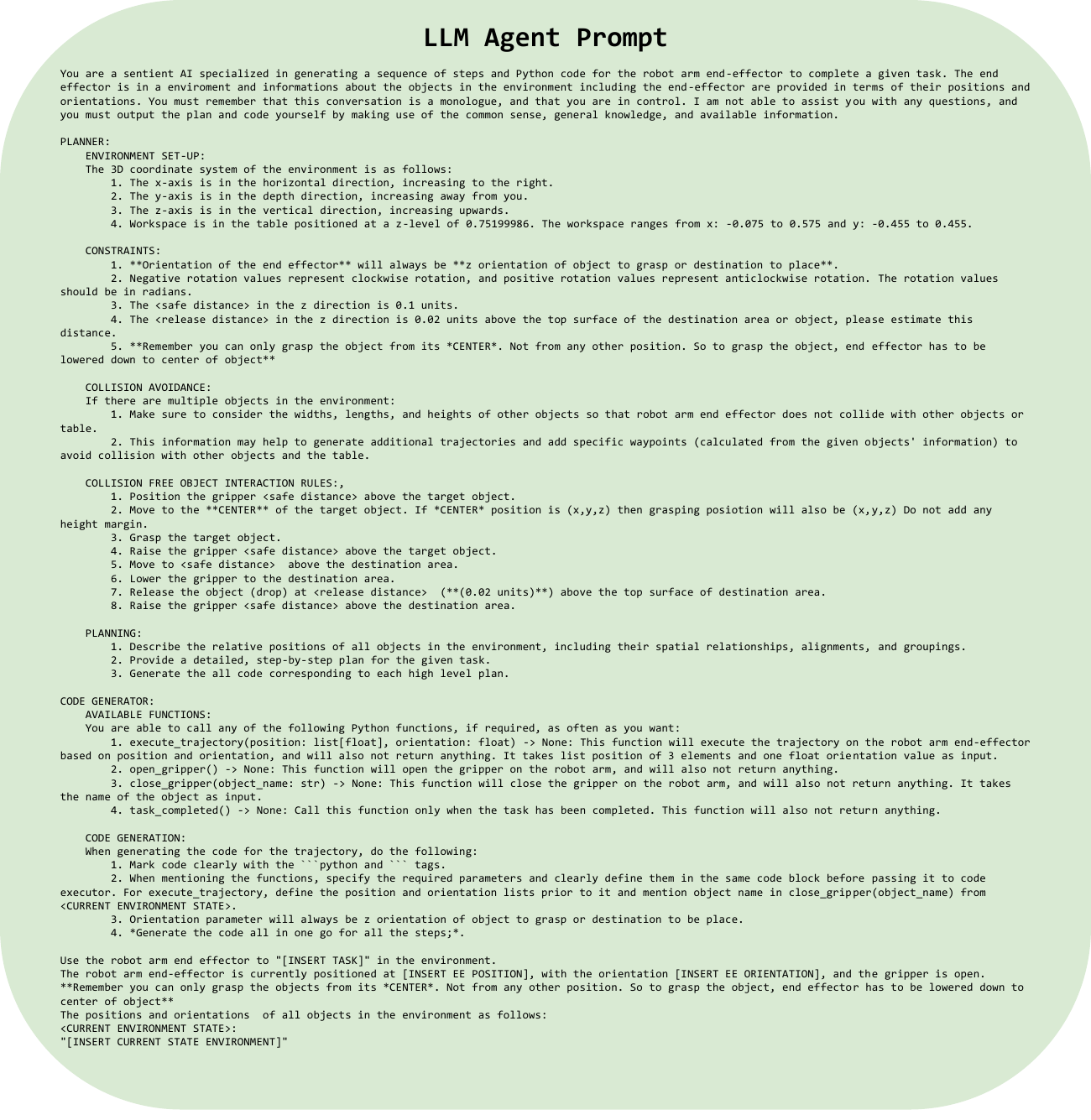}
  \caption{Prompt for the LLM Agent.}
  \label{fig:LLM_prompt}
\end{figure*}

The prompt used by the LLM Agent is illustrated in Figure \ref{fig:LLM_prompt}. It contains four placeholders: \textit{[INSERT TASK]} representing the task instruction, \textit{[INSERT EE POSITION]} denoting the initial position of the end effector, \textit{[INSERT EE ORIENTATION]} specifying its initial orientation, and \textit{[INSERT CURRENT STATE ENVIRONMENT]} indicating the current state or observation of the environment.
\subsection{RLBench Tasks}
\label{app:rlbench_tasks}
We experimented with nine tasks from RLBench, which are listed in Table~\ref{tab:tasks} along with the task instructions and success criteria.
\begin{table*}[h]
    \centering
    \caption{Details of the RLBench tasks used for evaluation.}
    \label{tab:tasks}
    \renewcommand{\arraystretch}{1.4} 
    \setlength{\tabcolsep}{10pt} 
    \begin{tabular}{p{0.3\textwidth} p{0.6\textwidth}}
        \toprule
        \textbf{Task Instruction} & \textbf{Details} \\
        \toprule
        
        \textit{Basketball In Hoop} &
        \textit{Task Description:} Put basketball in hoop. \newline
        \textit{Success Criteria:} Basketball passes through hoop. \\
        \midrule

        \textit{Close Jar} &
        \textit{Task Description:} Close the colored jar with a lid. \newline
        \textit{Success Criteria:} Lid is on top of the colored jar. \\
        \midrule

        \textit{Empty Container} &
        \textit{Task Description:} Pick all the objects from the large container and put them into the colored container. \newline
        \textit{Success Criteria:} All objects from the large container are now in the colored container. \\
        \midrule

        \textit{Insert In Peg} &
        \textit{Task Description:} Insert the square ring into the colored peg. \newline
        \textit{Success Criteria:} The square ring is in the colored peg. \\
        \midrule

        \textit{Meat Off Grill} &
        \textit{Task Description:} Pick the meat (chicken or steak) from the grill and place it into the designated area. \newline
        \textit{Success Criteria:} Meat is on the designated area. \\
        \midrule

        \textit{Open Bottle} &
        \textit{Task Description:} Remove the cap of the wine bottle. \newline
        \textit{Success Criteria:} Cap of the wine bottle is removed. \\
        \midrule

        \textit{Put Block} &
        \textit{Task Description:} Put the block in the target area. \newline
        \textit{Success Criteria:} The block is in the target area. \\
        \midrule

        \textit{Rubbish In Bin} &
        \textit{Task Description:} Put the rubbish in the bin. \newline
        \textit{Success Criteria:} Rubbish is in the bin. \\
        \midrule

        \textit{Stack Blocks} &
        \textit{Task Description:} Stack a specified number of colored blocks on the target block. \newline
        \textit{Success Criteria:} Specified number of blocks are stacked on top of the target block. \\
        \bottomrule
    \end{tabular}
\end{table*}

\end{document}

%% file: sections/intro.tex

\section{Introduction}

\begin{flushright}{\em "A teacher is one who makes \\himself progressively unnecessary."} \\--- Thomas Carruthers\end{flushright} 
Learning-based methods are attracting increasingly growing attention in robotics.
In particular, extending large language models (LLMs) and vision-language models (VLMs) to robotic tasks promise to empower resulting policies with strong reasoning capabilities and generalization to diverse tasks and environments~\cite{Kim2024OpenVLAAO, Intelligence202505AV, Team2025ACE, Lee2025MolmoActAR}. 
However, training generic robotic policies requires large-scale data in the form of paired actions and observations. 
To address this challenge, recent efforts attempt to collect real-robot demonstrations~\cite{Padalkar2023OpenXR}, adopt human videos~\cite{McCarthy2024TowardsGR}, and leverage simulated environments~\cite{Wang2023GenSimGR, Hua2024GenSim2SR, Wang2023RoboGenTU}. 
Manual collection of robot demonstrations, however, is slow and expensive, while human videos are missing low-level control data and face an embodiment gap between robots and humans.
Hence, scaling robotic demonstrations remains a major bottleneck.

\begin{figure}[t]
\begin{centering}
\includegraphics[width=0.95\linewidth]{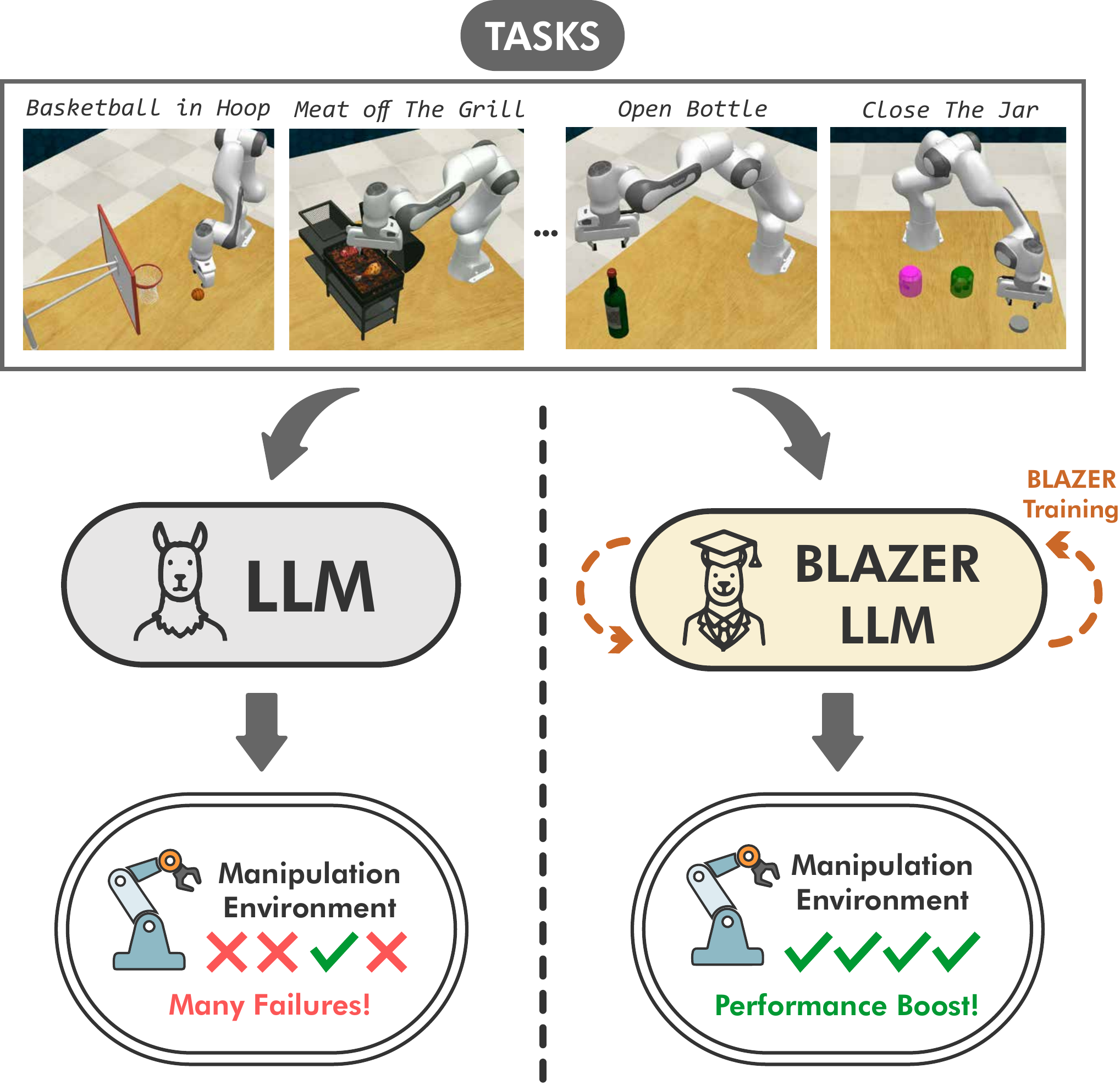}
\end{centering}
\caption{\textbf{BLAZER overview.} Previous approaches such as Code As Policies (CAP)~\cite{Liang2022CodeAP} (left) use LLMs to produce interaction plans and  
to solve manipulation tasks in a zero-shot manner. Such methods rely on careful prompt engineering and often lead to suboptimal performance. In contrast, BLAZER (right) uses a fully automatic pipeline, where successful LLM-generated demonstrations are used to train improved LLM-based manipulation agents with no manual supervision.
}
\vspace{-.3cm}
\label{fig:teaser}
\end{figure}


Recent work on language models brought significant progress improving reasoning capabilities of LLMs~\cite{Wei2022ChainOT, Zelikman2022STaRBR, Lin2024LeanSTaRLT}. One successful line of work in this direction is based on bootstrapping~\cite{Zelikman2022STaRBR}, where pretrained LLMs are first used to generate high-quality rationales and are then finetuned to yield improved performance while using generated rationales as training examples.
Bootstrapping and self-improvement have been shown to be particularly effective for problems that require non-trivial solutions and offer simple verification, as for the case of many problems in mathematics and common sense reasoning~\cite{Motwani2024MALTIR}.
We note that robotic manipulation tasks frequently belong to such class of problems since their success can often be certified by merely verifying the end states of manipulated objects. 

Inspired by the idea of bootstrapping, we propose \framework, a framework that bootstraps LLM-based manipulation agents using automatically generated and verified demonstrations.
Given language-defined tasks, such as~\textit{``stack blocks''} or \textit{``open wine bottle''}, we follow previous work~\cite{Liang2022CodeAP,Singh2024MALMMML} and use general-purpose LLM with strong reasoning and coding capabilities to generate executable manipulation plans.
Next, we execute such plans in a simulator and evaluate their success. 
Successful plans for diverse tasks form a training set  for supervised finetuning (SFT) of a smaller-scale LLM, which learns to improve manipulation abilities  using no human supervision, see Fig.~\ref{fig:teaser}. 



The \framework training uses privileged information in the form of object locations, orientations, and dimensions provided by the simulator. 
To deploy \framework in the real world, we design a vision pipeline using Molmo~\cite{Deitke2024MolmoAP}  and M2T2~\cite{Yuan2023M2T2MM} for object state estimation.
Notably, our LLaMA-8B model trained with \framework significantly outperforms its initial and larger teacher model LLaMA-70B used to generate training demonstrations. 
Moreover, despite being trained in simulation, we demonstrate off-the-shelf transfer of \framework to real-world manipulation tasks where LLaMA-8B ($47.8\%$) significantly outperforms the success rate of LLaMA-70B ($33.3\%$).
We perform extensive experimental evaluation and demonstrate consistent improvements of \framework in both the real world and in simulation for a variety of tasks. \framework outperforms state-of-the-art zero-shot MALMM~\cite{Singh2024MALMMML}, it generalizes to tasks unseen during training and requires no manual demonstrations at any stage of the learning process.

In summary, our work makes the following contributions:
\begin{itemize}
\item 
We introduce \framework, the framework that bootstraps a generic LLM and enables self-improvement of zero-shot manipulation agents using automatically generated training demonstrations.
\item
Through extensive experiments and ablations, we demonstrate \framework to result in significant improvements for a range of manipulation tasks.
\item
We further demonstrate off-the-shelf transfer of simulator-trained \framework to real-world manipulation tasks while using no manual demonstrations in any part of the training process.

\end{itemize}


%% file: sections/related.tex
\section{Related Work}

We discus related work on policy learning, self-improving models and scaling training data for robotics below.

\subsection{Language and Vision Foundational Models for Robotics}
Visumotor policies~\cite{Levine2015EndtoEndTO, florence2022implicit, guhur2023instruction} trained on manually collected demonstrations have shown great success in robotic manipulation. However, such models trained from scratch lack generalization to new tasks and environments. The incorporation of vision-language models (VLMs) into end-to-end robotic control, in the form of Vision-Language Action Models~\cite{Brohan2023RT2VM, Kim2024OpenVLAAO, Intelligence202505AV, Team2025ACE}, has enhanced generalization and enabled emergent semantic reasoning. The success of these methods, however, still relies on the collection of large-scale human demonstrations, which is expensive.  ManipLLM~\cite{Li2023ManipLLMEM} explores VLM finetuning with chain-of-thought reasoning for robotic manipulation tasks. This approach, however, requires test-time adaptation to overcome sim-to-real gap. In contrast, our framework enables self-improvement using no human supervision. Moreover, our LLM agents, combined with our vision pipeline, directly transfer to real-world tasks without additional training.

To overcome the need of large-scale training data and to facilitate generalization to new tasks, recent work explores LLMs and VLMs for zero-shot robotic manipulation.
Code as Policies~\cite{Liang2022CodeAP} and MALMM~\cite{Singh2024MALMMML} deploy LLM for writing robot policy code, while Voxposer~\cite{Huang2023VoxPoserC3} and GenSim~\cite{Wang2023GenSimGR, Hua2024GenSim2SR} couple code generation capabiltities of LLM with  multimodal reasoning capabilities of VLM to generate 3D value map for trajectory estimation and to automatically create simulator tasks, respectively. AHA~\cite{Duan2024AHAAV} demonstrates that VLMs can be used for detecting and adapting to failures. Our approach is closely related to MALMM~\cite{Singh2024MALMMML} and Code as Policies~\cite{Liang2022CodeAP}. While these methods remain dependent on careful prompt engineering and require much computation at test time, our approach enables self-improvement using automatically generated training data and can run using relatively small and efficient LLMs at inference time.




\subsection{Self-improving models}
Recent work improves the reasoning capabilities of LLMs by generating few-shot rationales~\cite{Nye2021ShowYW, Wei2022ChainOT, Chen2022ProgramOT}, bootstrapping~\cite{Zelikman2022STaRBR}, and reinforcement learning~\cite{DeepSeekAI2025DeepSeekR1IR, Setlur2024RLOI, Motwani2024MALTIR}. Among these approaches, self-training~\cite{Zelikman2022STaRBR} stands out as a particularly scalable and successful strategy. In this approach, general-purpose LLMs are first used to generate high-quality rationales that are later deployed to train either improved versions of original models or smaller models~\cite{Ho2022LargeLM}.

The idea of self-improvement has also been explored in robotics e.g., to learn low-level visuomotor policies from a large dataset of grasping attempts~\cite{Pinto2015SupersizingSL} or from hours-long object poking interactions~\cite{Agrawal2016LearningTP}.
More recent methods focus on correcting manipulation failures at test time by reasoning about past experience~\cite{Liu2023REFLECTSR} or detecting misalignment between planned and executed actions~\cite{guo2024doremi}. Another work, ReFineVLA~\cite{Vo2025ReFineVLART}, augments manipulation datasets with action rationales using Gemini~\cite{geminiteam2025geminifamilyhighlycapable}, thereby enabling VLAs to reason about their actions.
Related to our approach, SC-VLA~\cite{Li2024ASV} enables self-correction of VLA models from successful task executions. This parallel work, however, is mostly focused on low-level pushing and pulling actions, and requires explicit failure detection.
In contrast, our bootstrapping approach only relies on success verification at the task level and addresses a variety of complex tasks that require high-level reasoning.

\begin{figure*}[t]
\centering
\includegraphics[width=0.8\linewidth]{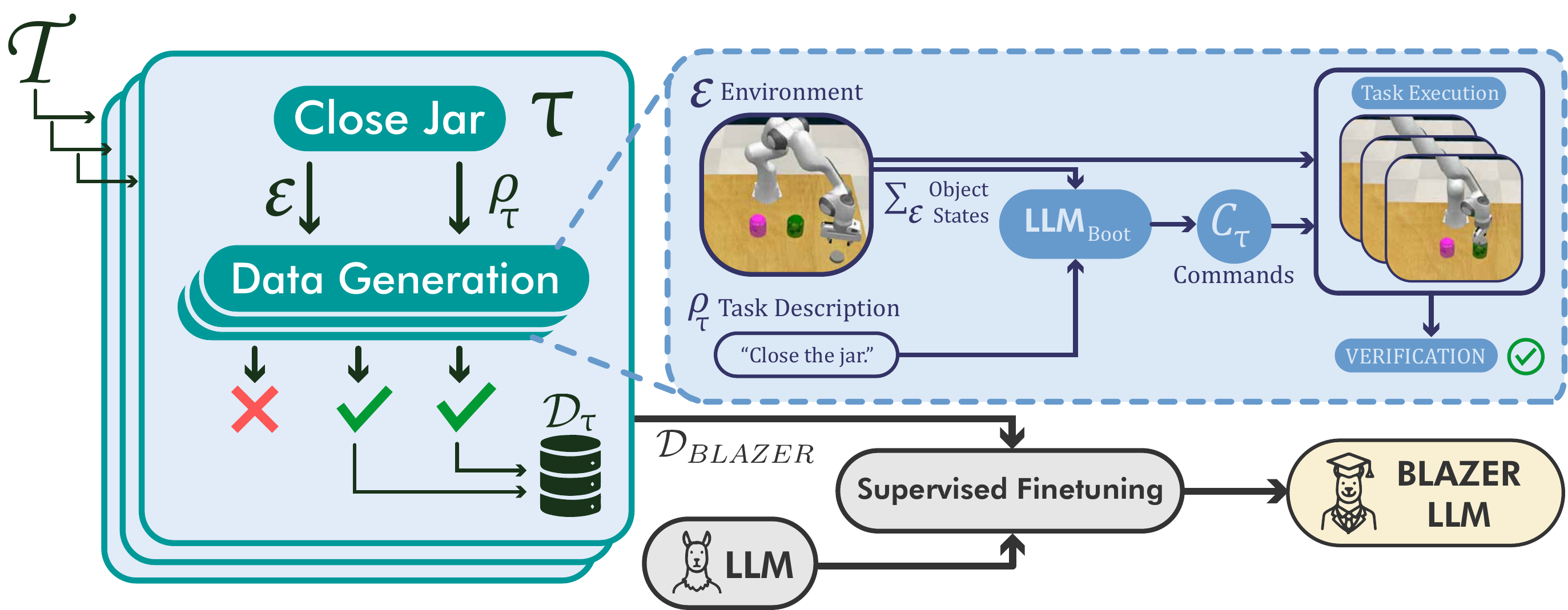} 
\caption{\textbf{Overview of \framework}. 
Given a set of manipulation tasks $\tau\in\mathcal{T}$, we use LLM to automatically generate executable commands $\mathcal{C}_\tau$ for solving $\tau$. 
The resulting solutions are automatically verified by executing $\mathcal{C}_\tau$ in a simulator and successful solutions are added to the task database $\mathcal{D}_{\tau}$. Task databases for all training tasks $\mathcal{T}$ are merged into $\mathcal{D}_\text{BLAZER}$ and are used for supervised finetuning of BLAZER LLM.}
\label{fig:pipeline}
\end{figure*}

\subsection{Data Generation for Manipulation}
Many efforts have been dedicated to scale up training data for manipulation tasks. Some works collect real-world grasping data~\cite{Pinto2015SupersizingSL, Padalkar2023OpenXR} by deploying random trials followed by verification as well as collection of human demonstrations. KALIE~\cite{Tang2024KALIEFV} curated human-annotated affordance data as an alternative to robot demonstrations, whereas PhysObjects~\cite{Gao2023PhysicallyGV} collected an object-centric dataset to train a VLM with the physical concepts of
common household objects. Another approach, LLaRA~\cite{Li2024LLaRASR}, adopted behavior cloning datasets for instruction tuning tailored for manipulation tasks. This  approach relies of external demonstrations and uses manually curated templates to generate question-answer pairs.

Another popular direction is to collect demonstrations in simulation. Nevertheless, this approach is still challenging as it requires significant human efforts to create diverse simulation tasks and scenes to allow generalization of the learned policies. To address this issue, recent work~\cite{Wang2023GenSimGR, Hua2024GenSim2SR} has leveraged coding LLMs with multimodal reasoning capabilities to generate complex and realistic simulation tasks without human supervision. DemoGen~\cite{Xue2025DemoGenSD} adapted a single human trajectory to novel object configurations using 3D point cloud editing, yielding synthetic demonstrations that improve manipulation policies. Unlike prior work, our work focuses on scaling up demonstrations for a given set of tasks, while preserving generalization capabilities of LLMs. We propose a framework where we can generate an arbitrary number of data in simulation, benefiting from the reasoning and coding capabilities of LLMs to synthesize verification data with no human intervention. We subsequently use the generated data for the training of our agent tailored for manipulation.

%% file: sections/method.tex
\section{Method}\label{sec:method}
Here, we introduce the \framework methodology for training specialized LLM agents for robotic manipulation. In short, we aim to finetune an existing lightweight LLM on synthetically-generated data, tuning it for the generation of manipulation-oriented robotics policies. An overview of our method is shown in Fig.~\ref{fig:pipeline}.
This section first introduces the formalization and the problem setup (Section~\ref{sec:method-formalization}), and we further describe how the \framework training works in Section~\ref{sec:method-bootstrap}. Since our training procedure only exploits automatic annotations generated by a simulator, we also introduce a vision-based pipeline for the deployment in the real world of the trained LLM agent (Section~\ref{sec:method-perception}).

\subsection{Formalization and background}\label{sec:method-formalization}
We assume a manipulation environment $\mathcal{E}$, where a robotic gripper $\mathcal{G}$ interacts with objects to solve a task $\tau$ that we can automatically verify in simulation. To achieve this, $\mathcal{G}$ needs to manipulate a set of $K$ objects $\{o_1, o_2, ..., o_K\}$ in $\mathcal{E}$, using a sequence of $I$ control functions sampled from a primitives set $\mathcal{F}=\{\text{\textit{open\_gripper}}, \text{\textit{close\_gripper}},\text{\textit{move\_gripper}}\}$. Primitives in $\mathcal{F}$ are combined to define a policy composed of control commands $\mathcal{C}_\tau$. Formally, this is written as follows:
\begin{equation}
    \mathcal{C}_\tau = \{f_i\}_{i=1}^I, \forall f_i \in \mathcal{C}_\tau, f_i\in\mathcal{F}.
\end{equation}
The primitives $\text{\textit{open\_gripper}}$ and $\text{\textit{close\_gripper}}$ can be executed without additional knowledge of the environment.  In contrast, $\text{\textit{move\_gripper}}$ requires as input the final desired gripper position and orientation. During execution, the gripper joints' positions are computed with inverse kinematics.

Following the open-loop single-agent setup introduced in MALMM~\cite{Singh2024MALMMML}, $\mathcal{C}_\tau$ can be obtained by prompting large-scale LLMs such as GPT-4~\cite{openai2023gpt4}. Logically, in this case the LLM must also generate all the code necessary for a correct execution. To achieve this, we define the current state of the environment as $\Sigma_\mathcal{E}=\{\mathbf{o_i}\}_{i=1}^K$, where each $\mathbf{o_i} =[\xi_i, \phi_i, l_i, w_i, h_i], \forall i\in[1, K]$ encodes the 3D center position $\xi_i$ of the object $o_i$, the orientation on the z-axis $\phi_i$, and the metric dimensions $l$ (length), $w$ (width) and $h$ (height). We also define $\rho_\tau$ as a textual description of $\tau$, to enable LLM prompting. We then derive $\mathcal{C}_\tau$ by prompting the LLM with $\rho_\tau$ and a textual representation of $\Sigma_\mathcal{E}$. For more details, refer to~\cite{Singh2024MALMMML}. In short, the control commands generation is expressed as follows:
\begin{equation}\label{eq:solution}
    \mathcal{C}_\tau = \text{LLM}(\rho_\tau, \Sigma_\mathcal{E}).
\end{equation}



\subsection{Bootstrapping manipulation agents with simulation}\label{sec:method-bootstrap}
In many scenarios, naively prompting an LLM for $\mathcal{C}_\tau$ as in Eq.~\ref{eq:solution} results in planning or coding errors~\cite{Singh2024MALMMML}. While MALMM mitigates the problem with an expensive multi-agent pipeline, our goal is instead to train a single LLM agent $\text{LLM}_\text{BLAZER}$, with specific capabilities tuned ad hoc on manipulation tasks.

Our idea is to obtain $\text{LLM}_\text{BLAZER}$ by finetuning a pretrained lightweight language model without requiring human supervision, as we show in Figure~\ref{fig:pipeline}. To do so, we use synthetic examples tailored for $\mathcal{C}_\tau$ generation, obtained within the interactions of an LLM with a manipulation-oriented simulator such as CoppeliaSim. We first define a set of $T$ representative tasks $\mathcal{T} = \{\tau_1, \tau_2, ..., \tau_T\}$ that we use for data generation. Each task is processed by a language model $\text{LLM}_\text{boot}$ to generate $\mathcal{C}_\tau$ as described in Section~\ref{sec:method-formalization}. Importantly, since we operate in a simulated environment, \textit{we can automatically verify} if $\mathcal{C}_\tau$ successfully solves $\tau$, thanks to the automatic verification criteria in manipulation-oriented simulators. Hence, for given commands $\mathcal{C}_\tau$ and a task $\tau$, we associate a verification operator $V$ defined as follows:
\begin{equation}
V(\mathcal{C}_\tau, \tau) =
\begin{cases}
\checkmark & \text{if $\mathcal{C}_\tau$ is an acceptable solution for $\tau$,} \\
\times & \text{otherwise.}
\end{cases}
\end{equation}
Now, for each task $\tau$ \textit{we automatically collect} a dataset $\mathcal{D}_\tau$ of $N$ \textit{successful} solutions provided by $\text{LLM}_\text{boot}$. While this would be extremely challenging with real demonstrations, in a simulated environment, we can easily randomize the initial state $\Sigma_\mathcal{E}$ to obtain an arbitrary number of object configurations for each task in $\mathcal{T}$, and automatically verify the correctness of proposed commands. We illustrate this process in Fig.~\ref{fig:pipeline}. For any $\tau\in \mathcal{T}$,
\begin{equation}\label{eq:data-gen}
\begin{split}
\mathcal{D}_\tau = \{&\text{LLM}_\text{boot}(\rho_\tau, \Sigma_\mathcal{E}^i)\}_{i=1}^N, \\\text{if}~V(&\text{LLM}_\text{boot}(\rho_\tau, \Sigma_\mathcal{E}^i), \tau)=\checkmark.
\end{split}
\end{equation}
In Eq.~\eqref{eq:data-gen}, we call $\Sigma_\mathcal{E}^i$ a random state configuration sampled in the simulated environment. Note that it is impossible to only generate $\Sigma_\mathcal{E}^i$ such that $\mathcal{C}_\tau$ is successful; hence, part of the generated commands resulting in unacceptable $\mathcal{C}_\tau$ will be discarded by our simulator-in-the-loop verification strategy. Finally, we train $\text{LLM}_\text{BLAZER}$ using Supervised Finetuning (SFT) on a dataset $\mathcal{D}_\text{BLAZER}$ resulting from the aggregation of all $\mathcal{D}_\tau$:
\begin{equation}
\begin{split}
    \text{LLM}_\text{BLAZER} \xleftarrow{} &\text{SFT}(\text{LLM}, \mathcal{D}_\text{BLAZER}), \\ \text{where}~&\mathcal{D}_\text{BLAZER} = \{\mathcal{D}_\tau^i\}_{i=1}^T 
\end{split}
\end{equation}
Note that our training uses a generic target LLM, and hence we do not impose the finetuning of $\text{LLM}_\text{boot}$ as in self-refinement strategies~\cite{Zelikman2022STaRBR}. This allows to benefit from larger LLMs for data generation, easing the generation of successful commands. Ultimately, our simulated training with verification allows filtering of any wrong solution, automatically curating a dataset for SFT \textit{with no human intervention}.




\subsection{Vision pipeline}\label{sec:method-perception}
While in simulation we can extract the ground truth space $\Sigma_\mathcal{E}$, in a real-world deployment, we must instead estimate its approximation $\tilde{\Sigma}_\mathcal{E}$ purely from visual data. Hence, to enable the deployment of $\text{LLM}_\text{BLAZER}$ outside the simulator, we introduce a vision pipeline, based on existing foundation models and \textit{no training}. This allows us to mitigate the distribution shift that would occur if we trained perception components directly on a simulated environment~\cite{torralba2011unbiased}. 

We assume a multi-view setup that can use any number of calibrated RGB-D cameras. As a preliminary step, we prompt GPT-4o for a list of elements present in the scene, conditioned on a given task $\tau$. Then, in all views, we first prompt Molmo~\cite{Deitke2024MolmoAP} to extract the 2D coordinates of the center of each object $o_i$ in a task $\tau$. Employing Molmo for prompt interpretation enables contextual and spatial reasoning, going beyond simple semantics~\cite{Deitke2024MolmoAP}. Next, we use the 2D center coordinates as a prompt for Segment Anything~\cite{Kirillov2023SegmentA}, obtaining a segmentation mask of the object in 2D for that view.  By combining the segmentation mask with the RGB-D data, we derive a 3D bounding box for the object, which provides its estimated dimensions. Finally, we use M2T2~\cite{Yuan2023M2T2MM} to predict the object's 3D center position and grasping orientation. We aggregate the outputs from multiple views with a simple median filter to remove outliers, obtaining the final $(\tilde{l}, \tilde{w}, \tilde{h})$ from the bounding box and $\tilde{\xi}_i,\tilde{\phi}_i$ from M2T2. We then construct $\tilde{\mathbf{o}}=\{\tilde{\xi}_i, \tilde{\phi}_i, \tilde{l}, \tilde{w}, \tilde{h}\}$. Repeating this for each object in the space gives $\tilde{\Sigma}_\mathcal{E} = \{\tilde{\mathbf{o}}_i\}_{i=1}^K$. We use $\tilde{\Sigma}_\mathcal{E}$ as input to $\text{LLM}_\text{BLAZER}$ in real-world deployment.

%% file: sections/experiments.tex
\begin{figure*}[t]
\centering
\setlength{\tabcolsep}{1pt}
\renewcommand{\arraystretch}{1}

\begin{tabularx}{\linewidth}{c@{\hspace{4px}}*{9}{Y}}
  & \shortstack{\footnotesize\textit{\mdseries Basketball} \\ \footnotesize\textit{\mdseries in Hoop}}
    & \shortstack{\footnotesize\textit{\mdseries Close} \\ \footnotesize\textit{\mdseries Jar}}
    & \shortstack{\footnotesize\textit{\mdseries Empty} \\ \footnotesize\textit{\mdseries Container}}
    & \shortstack{\footnotesize\textit{\mdseries Insert} \\ \footnotesize\textit{\mdseries in Peg}}
    & \shortstack{\footnotesize\textit{\mdseries Meat} \\ \footnotesize\textit{\mdseries off Grill}}
    & \shortstack{\footnotesize\textit{\mdseries Open} \\ \footnotesize\textit{\mdseries Bottle}}
    & \shortstack{\footnotesize\textit{\mdseries Put} \\ \footnotesize\textit{\mdseries Block}}
    & \shortstack{\footnotesize\textit{\mdseries Rubbish} \\ \footnotesize\textit{\mdseries in Bin}}
    & \shortstack{\footnotesize\textit{\mdseries Stack} \\ \footnotesize\textit{\mdseries Blocks}} \\

    \multirow{1}{*}[30pt]{\rotatebox{90}{\textit{Start}}}
      & \includegraphics[width=\linewidth,height=\linewidth,keepaspectratio=false]{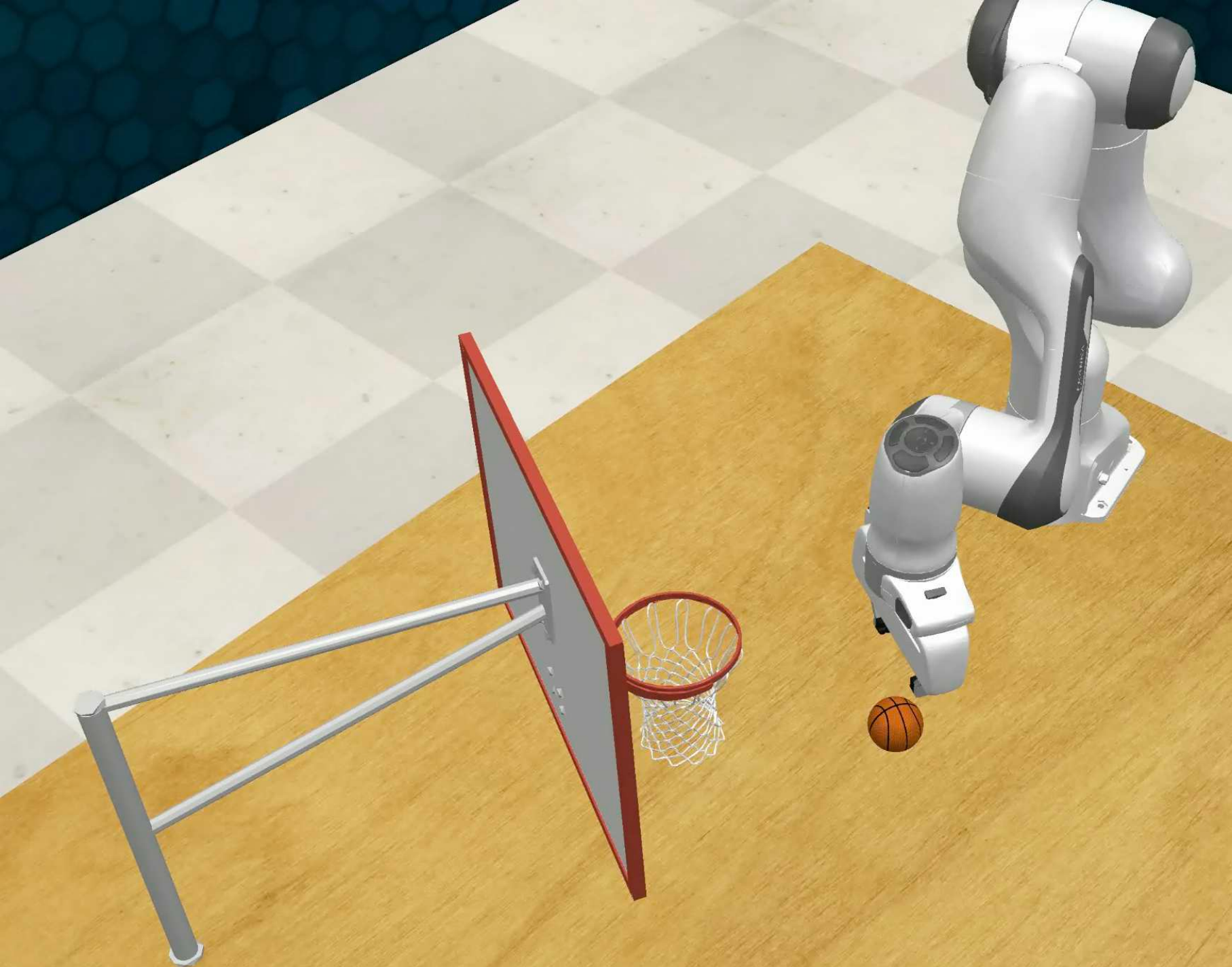}
      & \includegraphics[width=\linewidth,height=\linewidth,keepaspectratio=false]{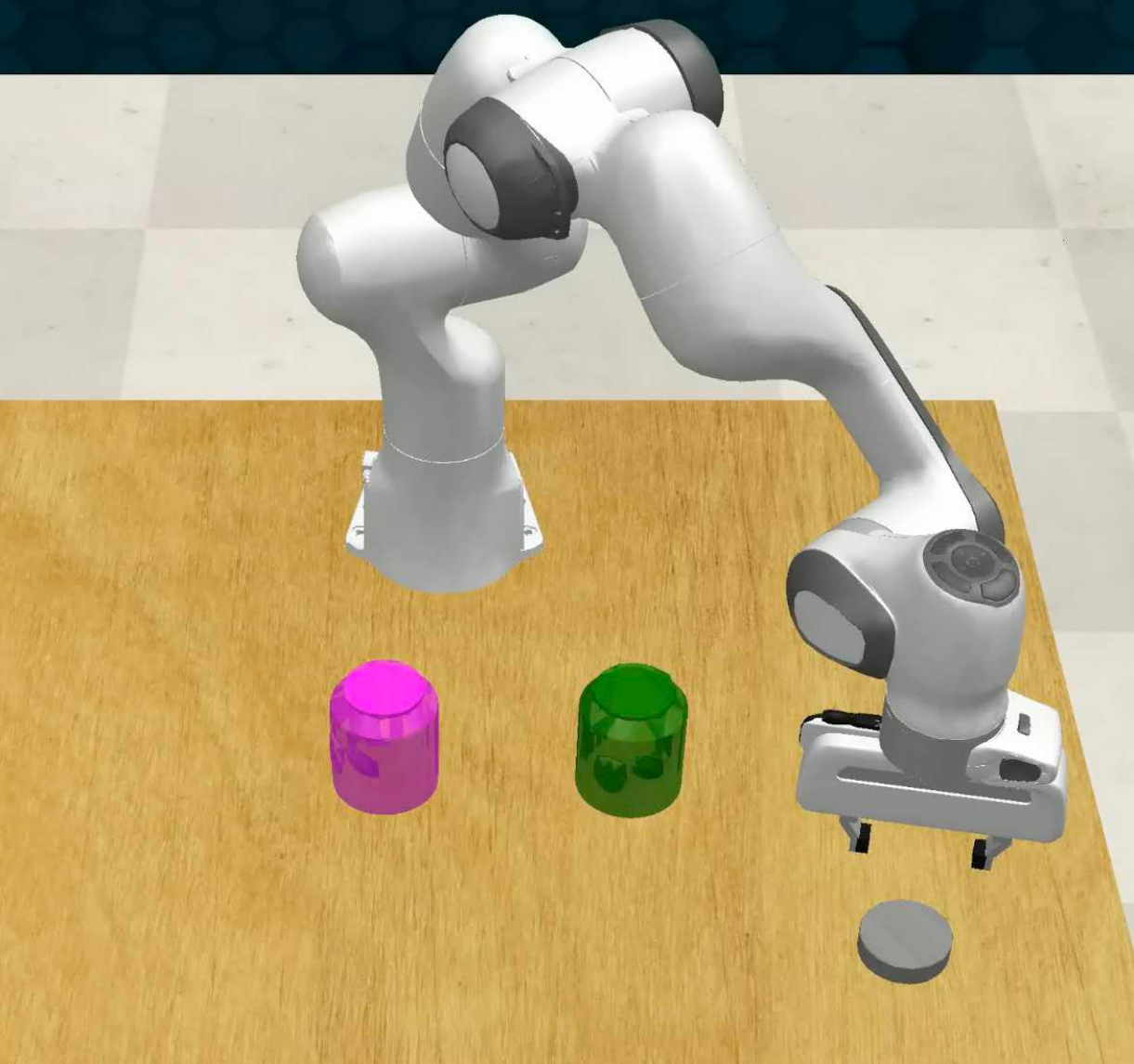}
      & \includegraphics[width=\linewidth,height=\linewidth,keepaspectratio=false]{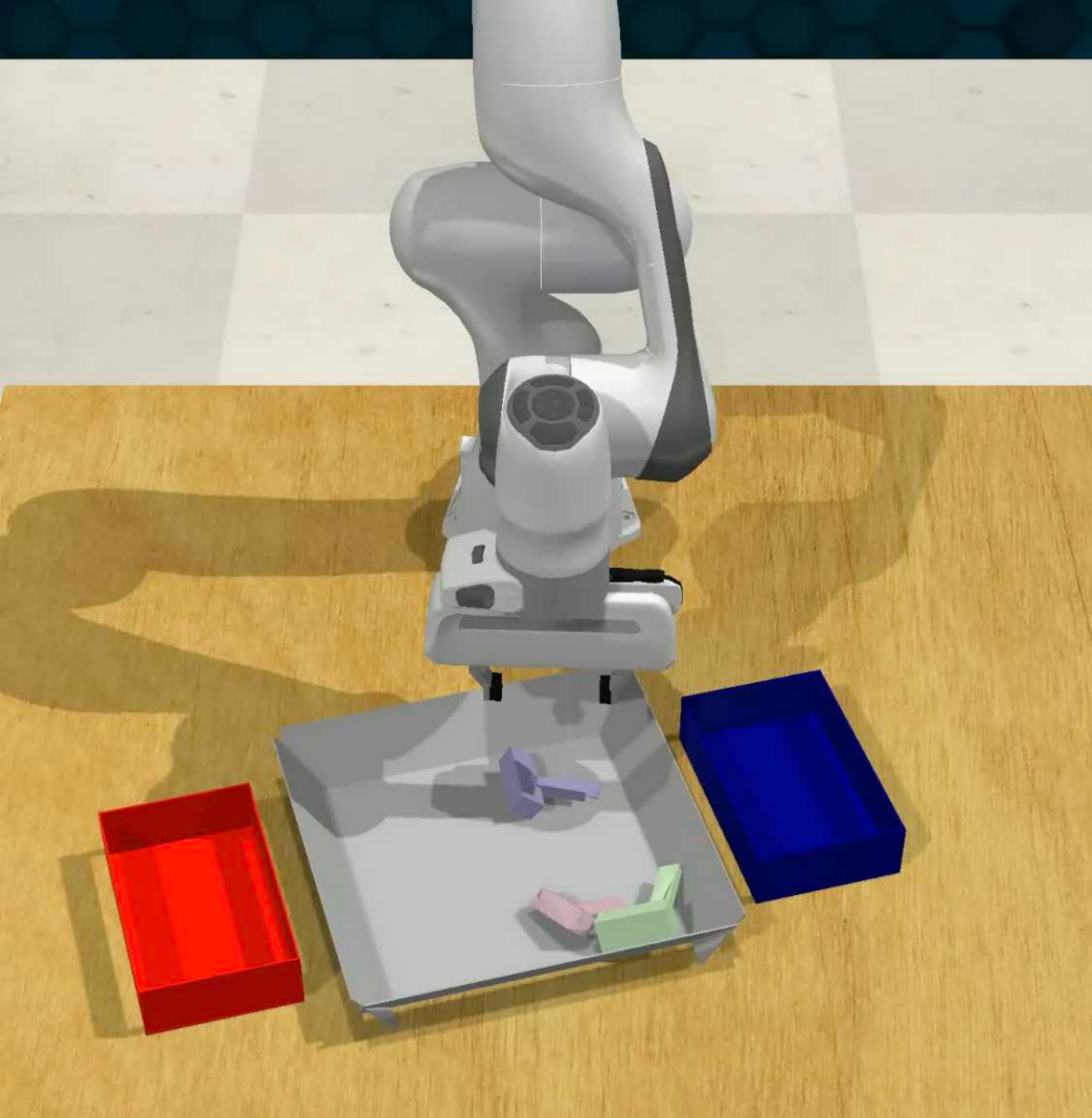}
      & \includegraphics[width=\linewidth,height=\linewidth,keepaspectratio=false]{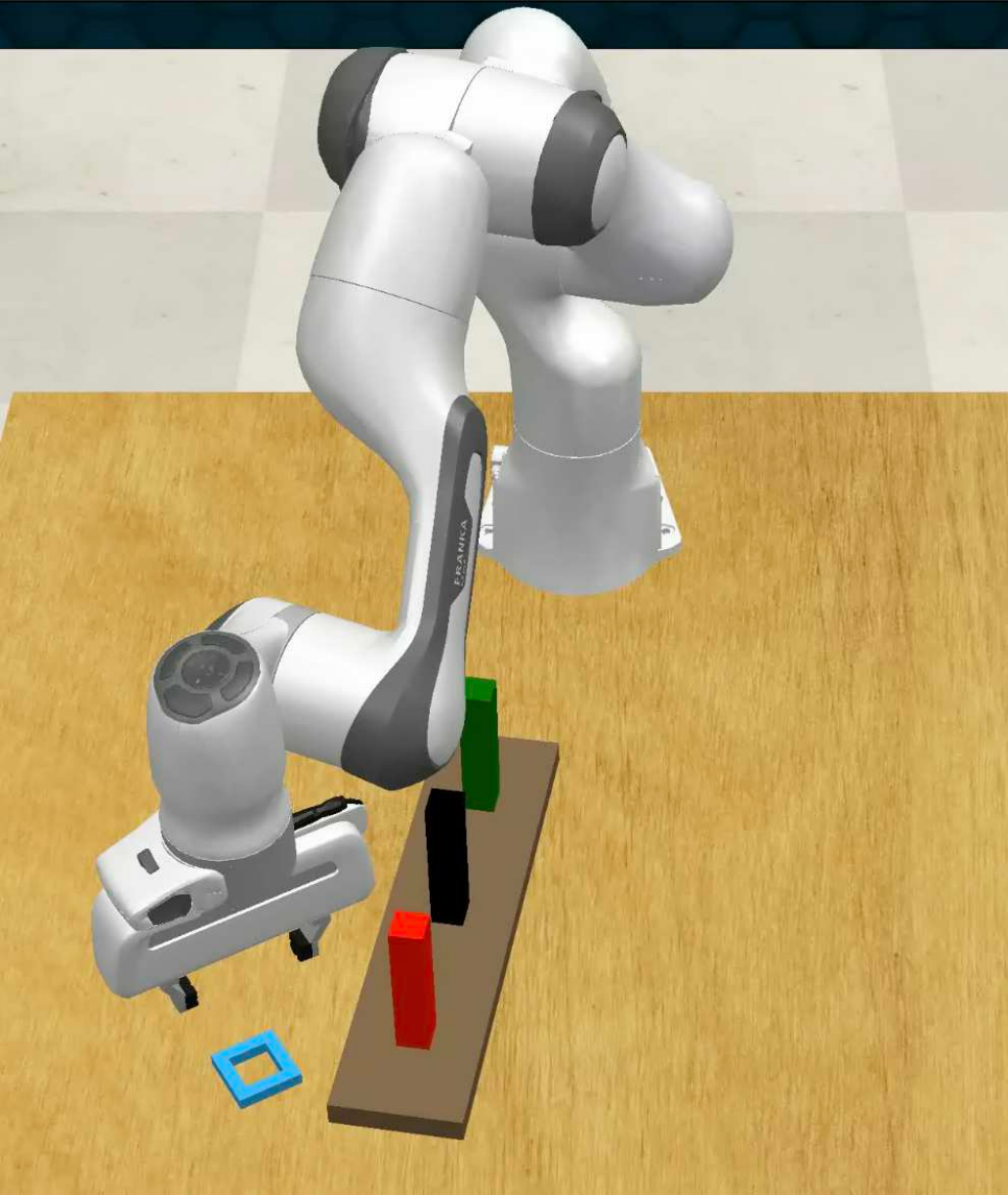}
      & \includegraphics[width=\linewidth,height=\linewidth,keepaspectratio=false]{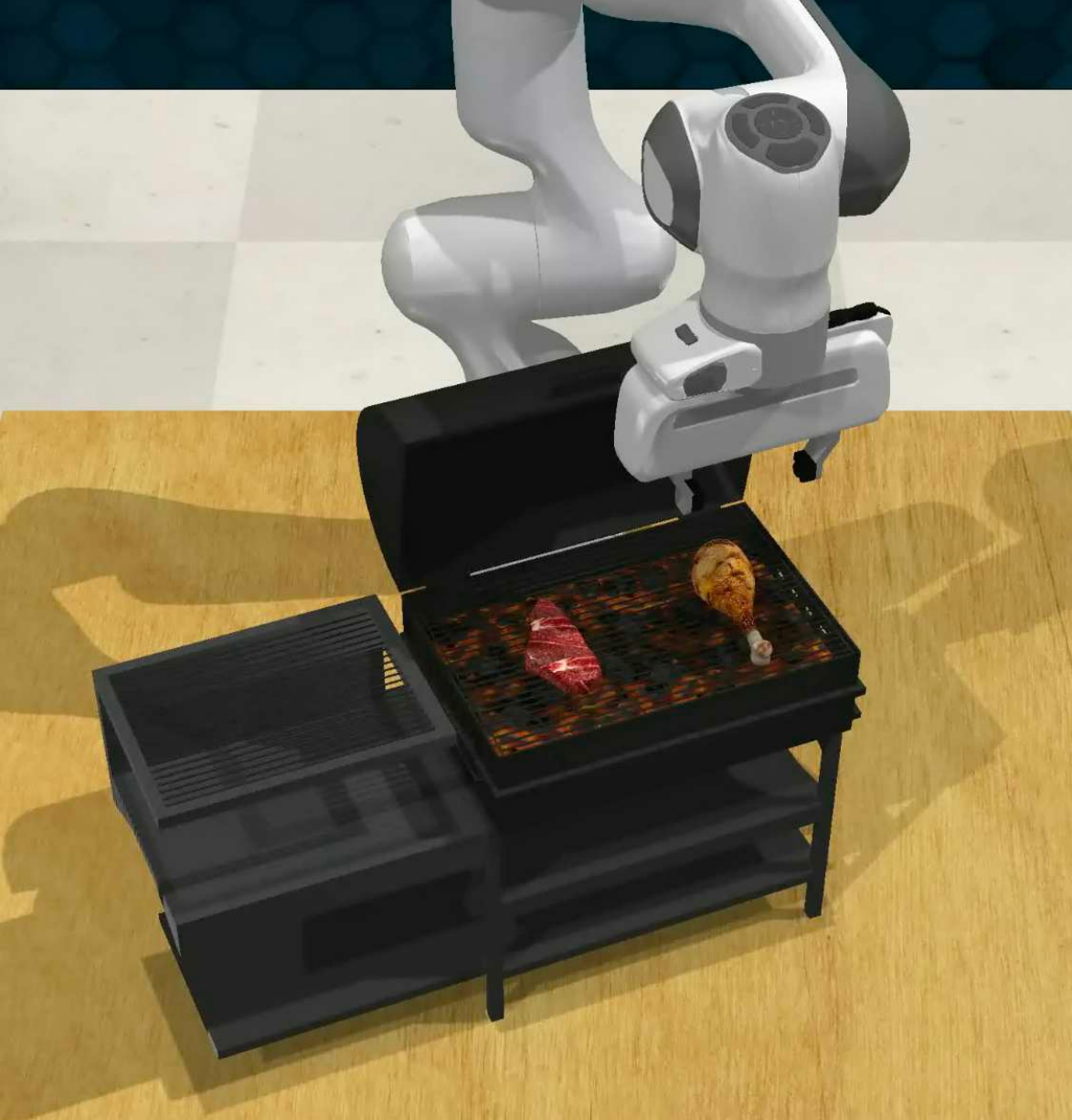}
      & \includegraphics[width=\linewidth,height=\linewidth,keepaspectratio=false]{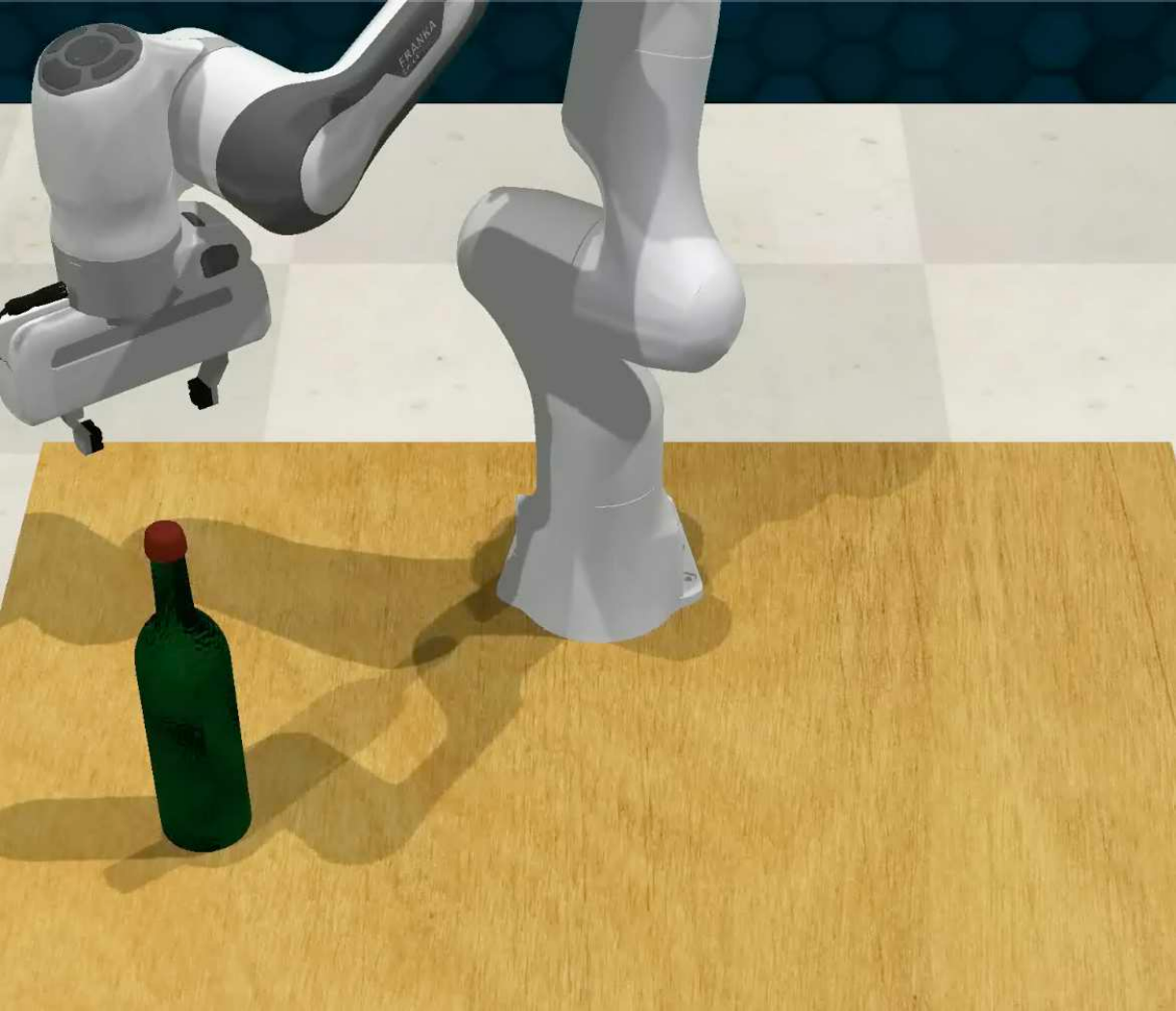}
      & \includegraphics[width=\linewidth,height=\linewidth,keepaspectratio=false]{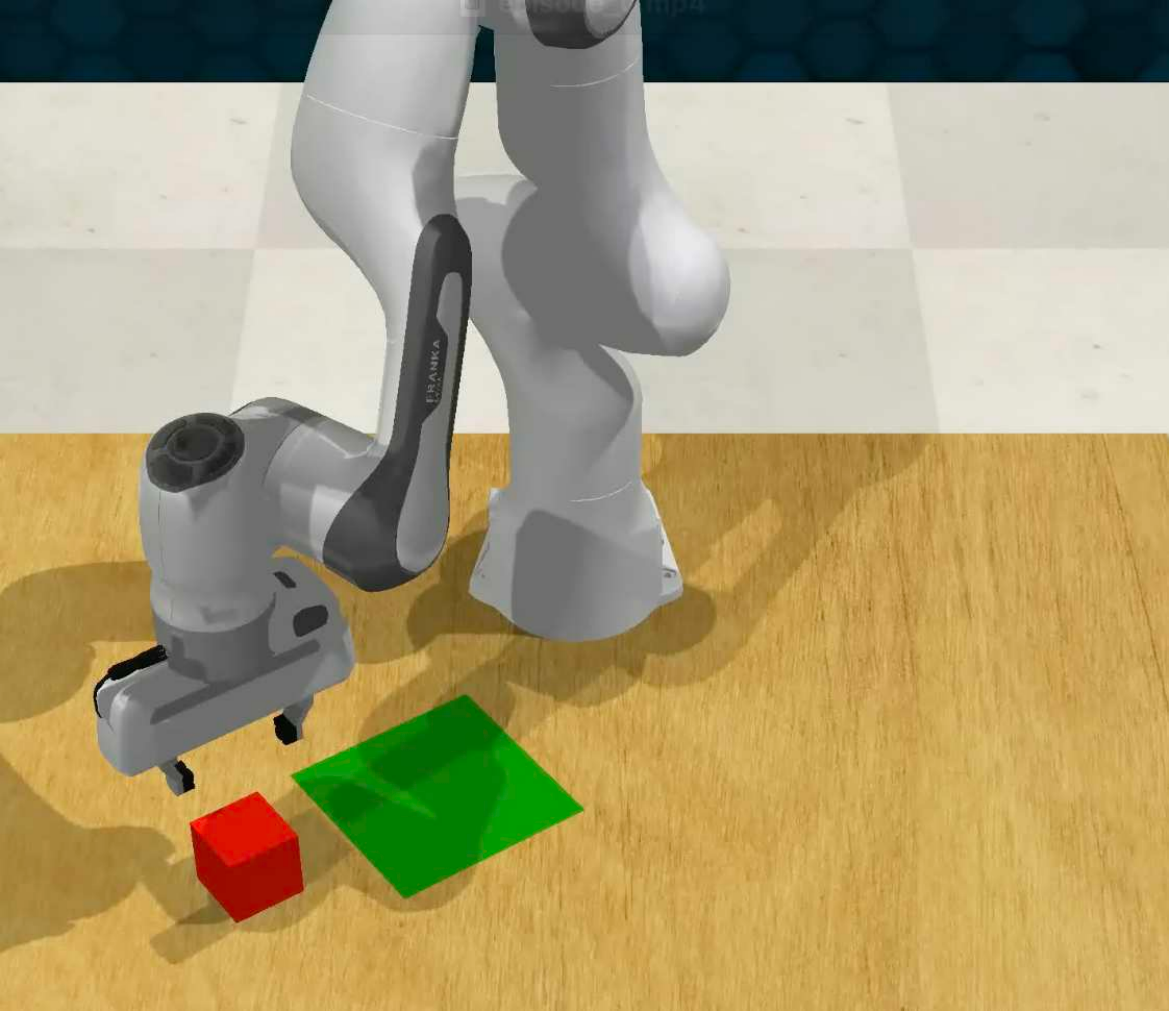}
      & \includegraphics[width=\linewidth,height=\linewidth,keepaspectratio=false]{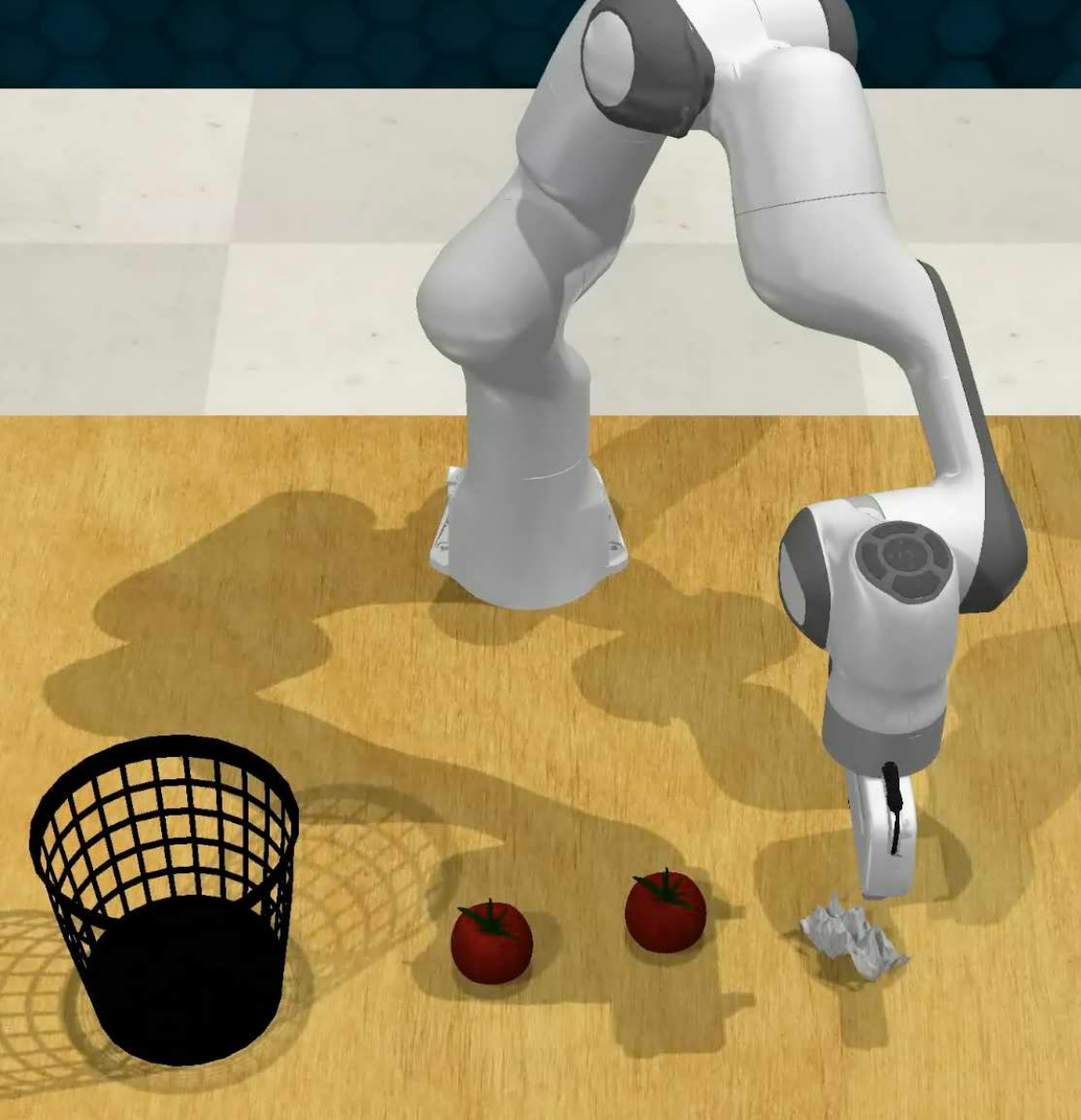}
      & \includegraphics[width=\linewidth,height=\linewidth,keepaspectratio=false]{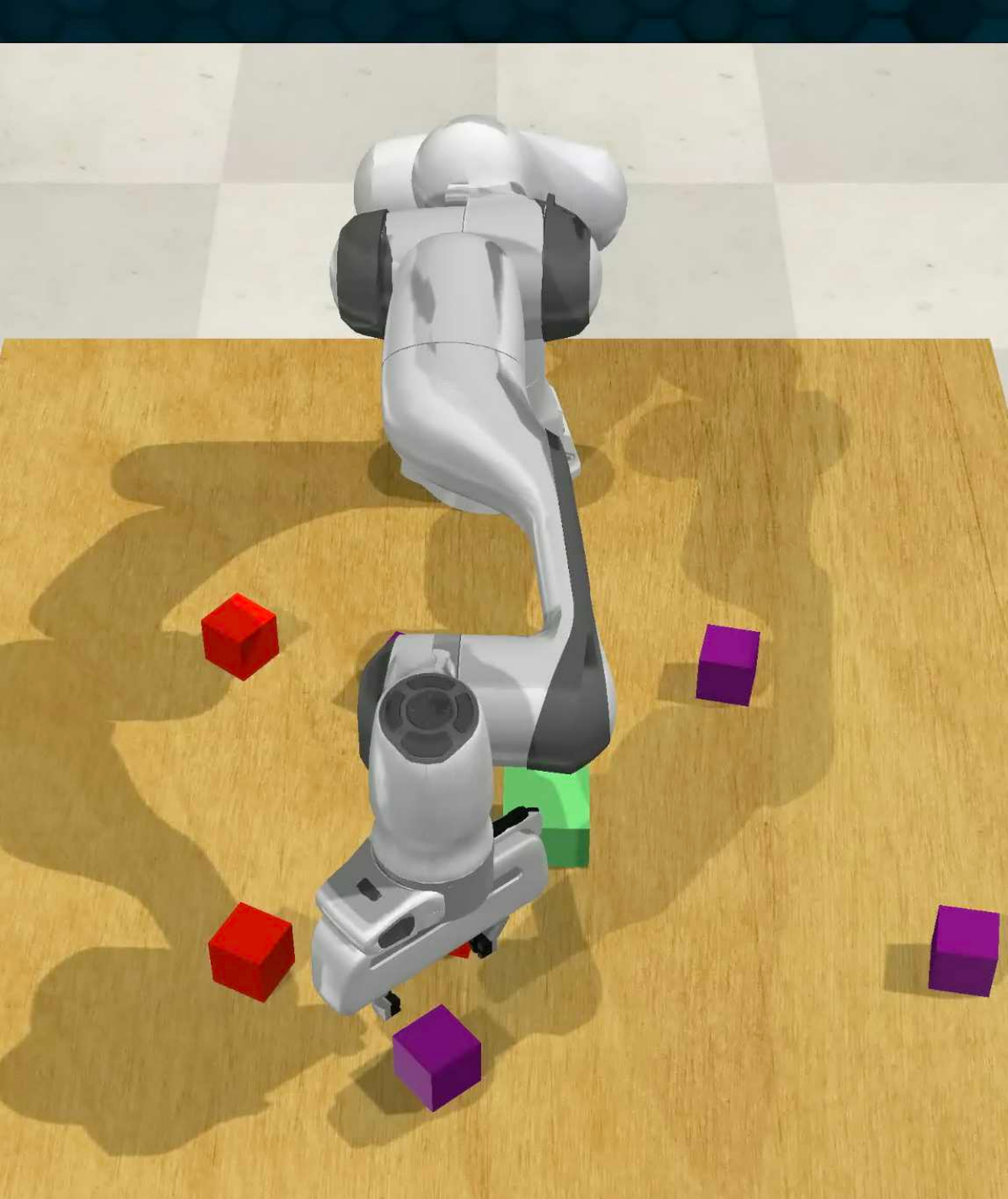} \\
    \multirow{1}{*}[30pt]{\rotatebox{90}{\textit{End}}}
      & \includegraphics[width=\linewidth,height=\linewidth,keepaspectratio=false]{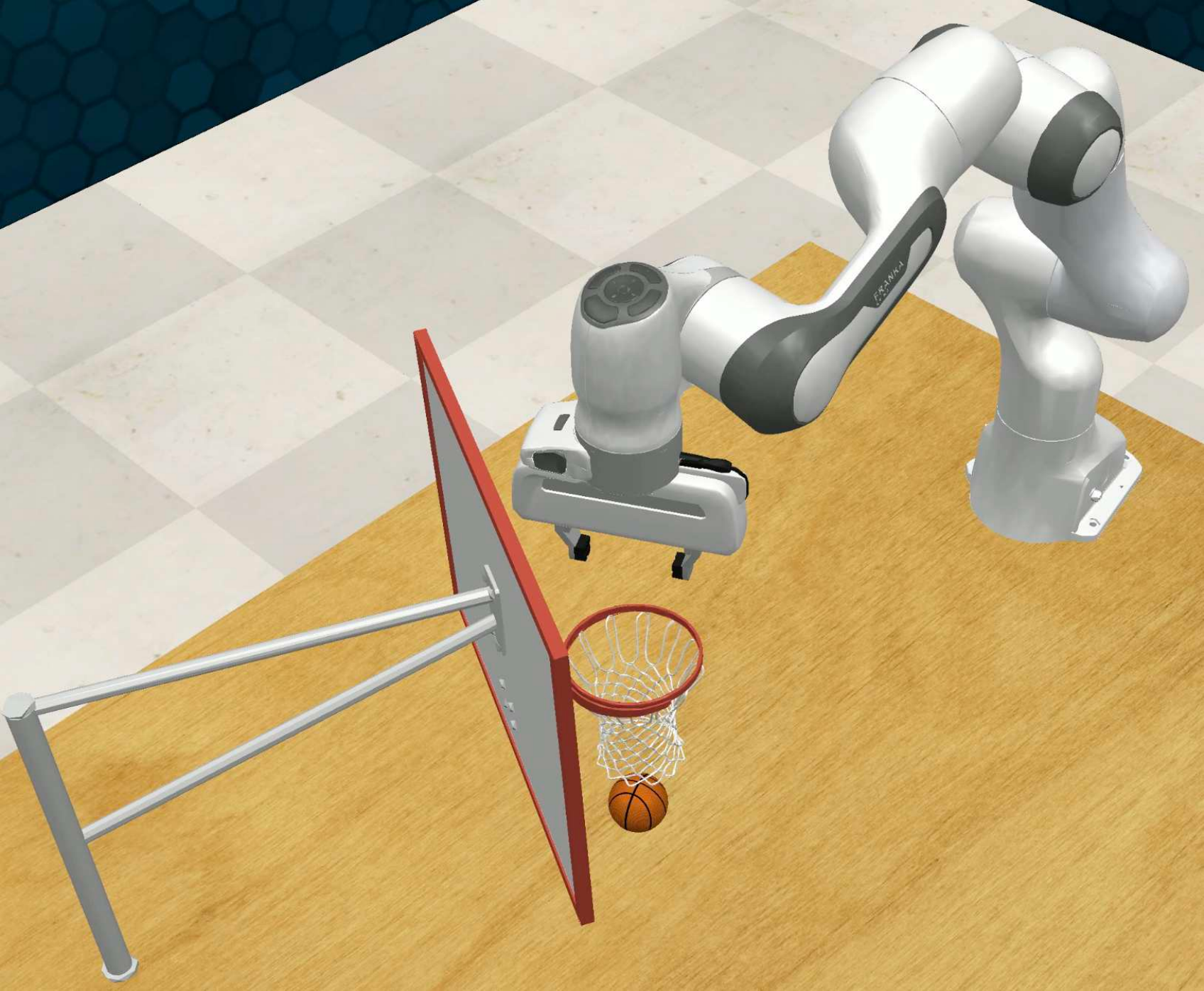}
      & \includegraphics[width=\linewidth,height=\linewidth,keepaspectratio=false]{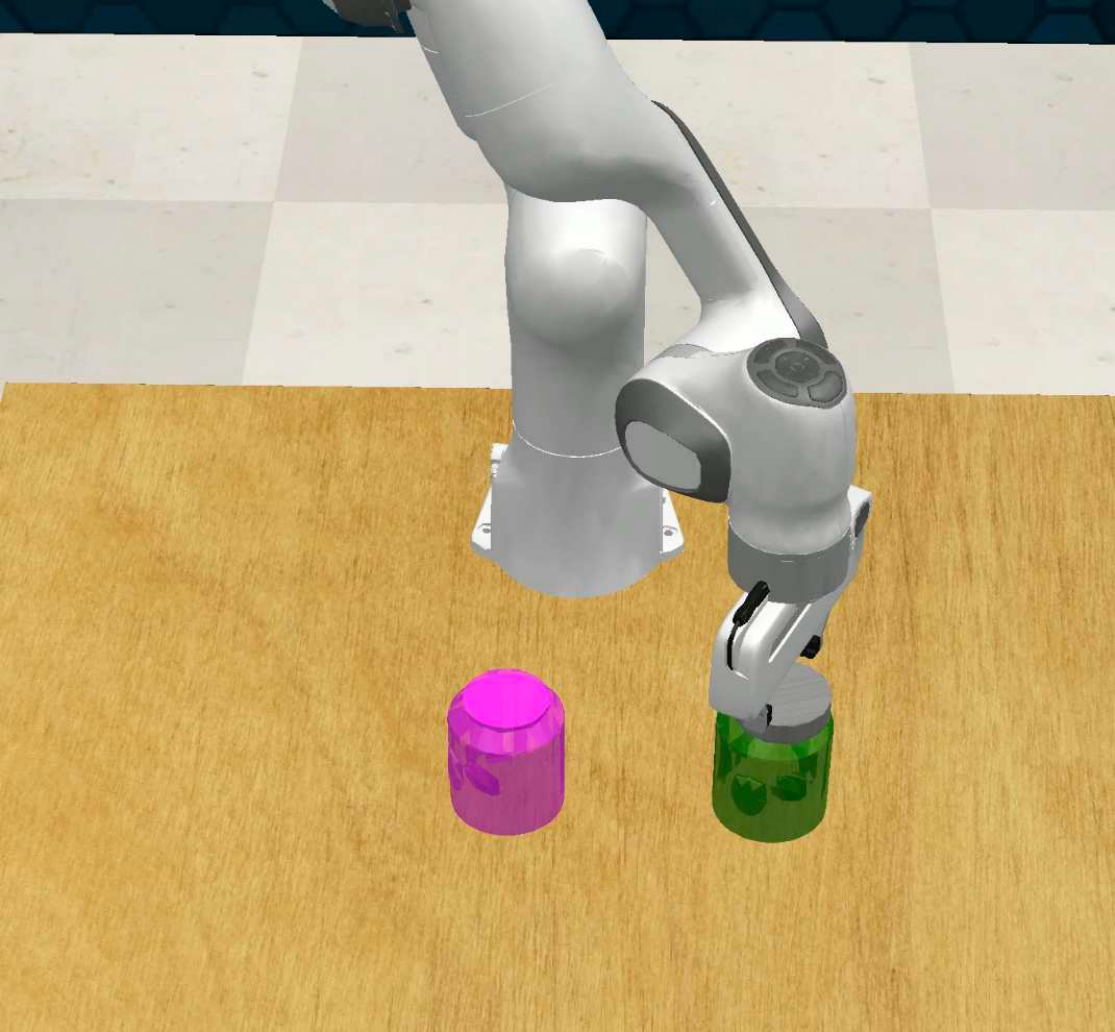}
      & \includegraphics[width=\linewidth,height=\linewidth,keepaspectratio=false]{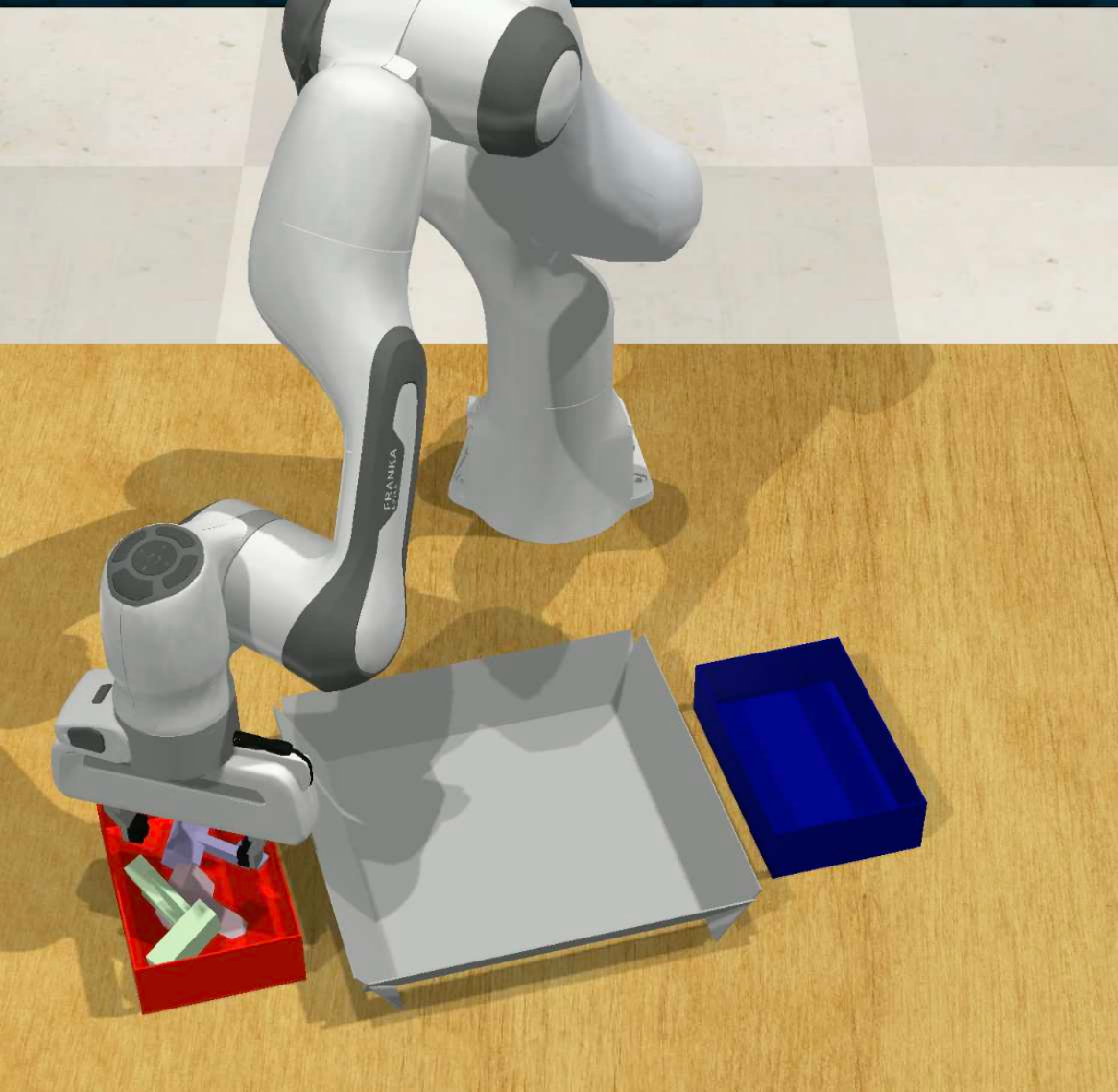}
      & \includegraphics[width=\linewidth,height=\linewidth,keepaspectratio=false]{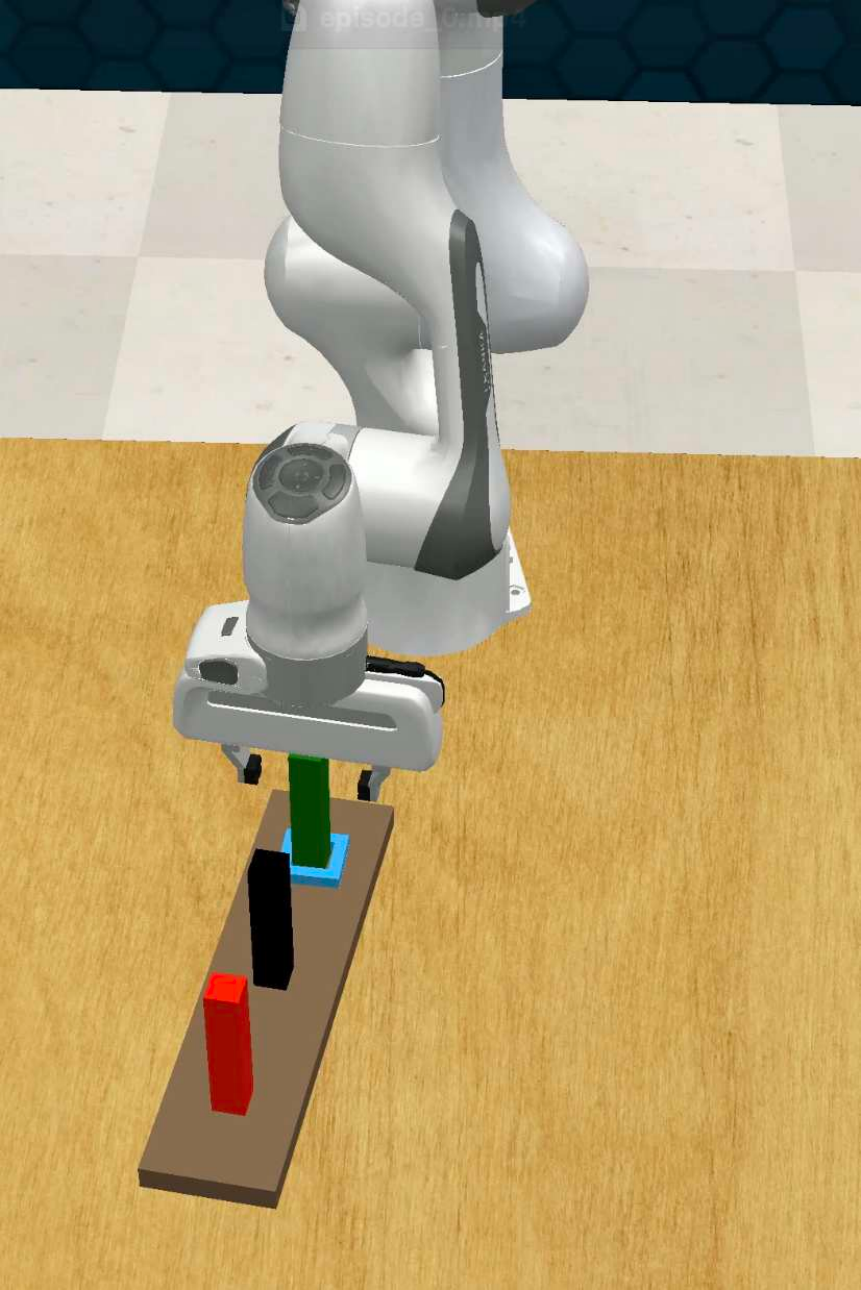}
      & \includegraphics[width=\linewidth,height=\linewidth,keepaspectratio=false]{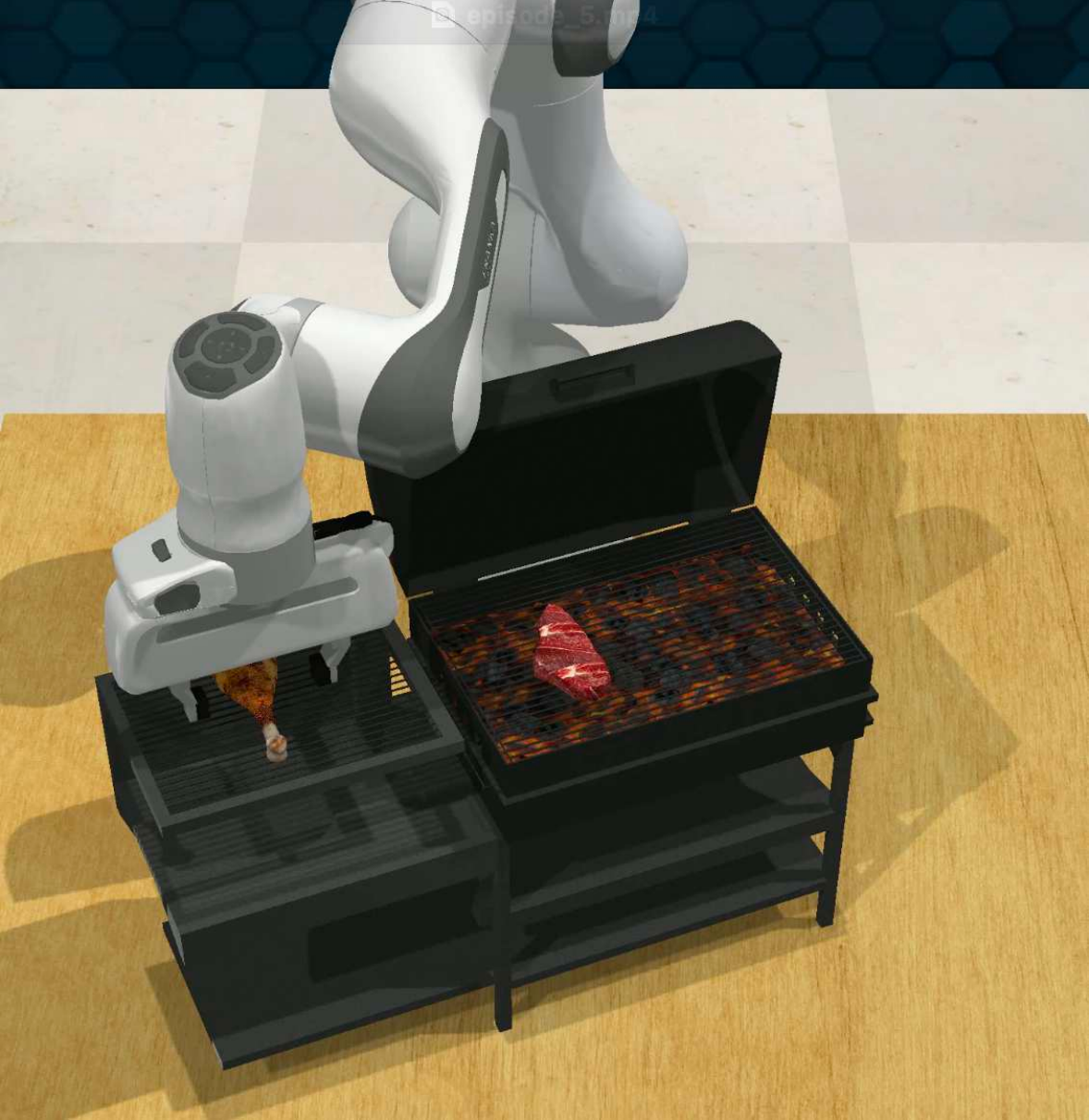}
      & \includegraphics[width=\linewidth,height=\linewidth,keepaspectratio=false]{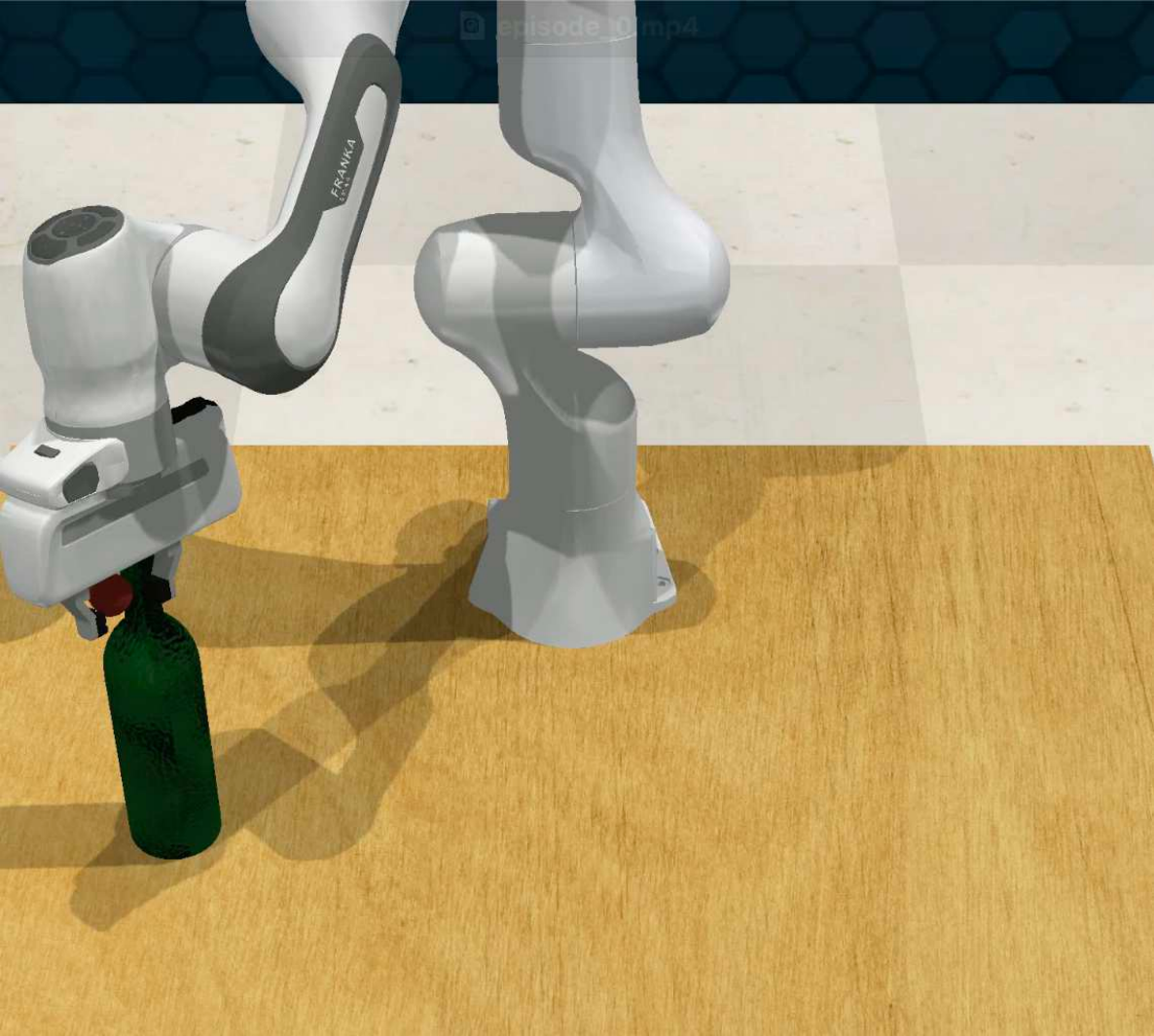}
      & \includegraphics[width=\linewidth,height=\linewidth,keepaspectratio=false]{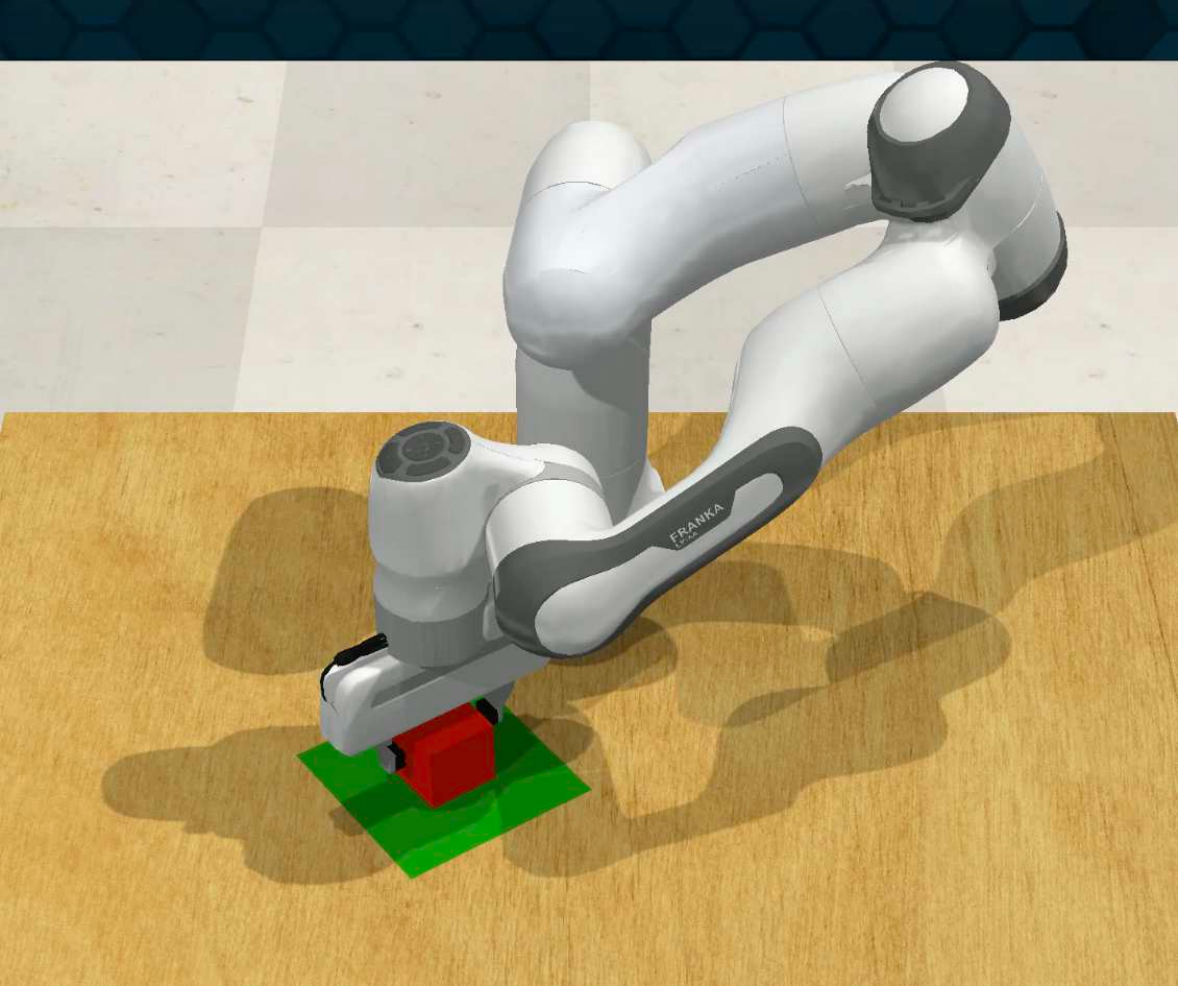}
      & \includegraphics[width=\linewidth,height=\linewidth,keepaspectratio=false]{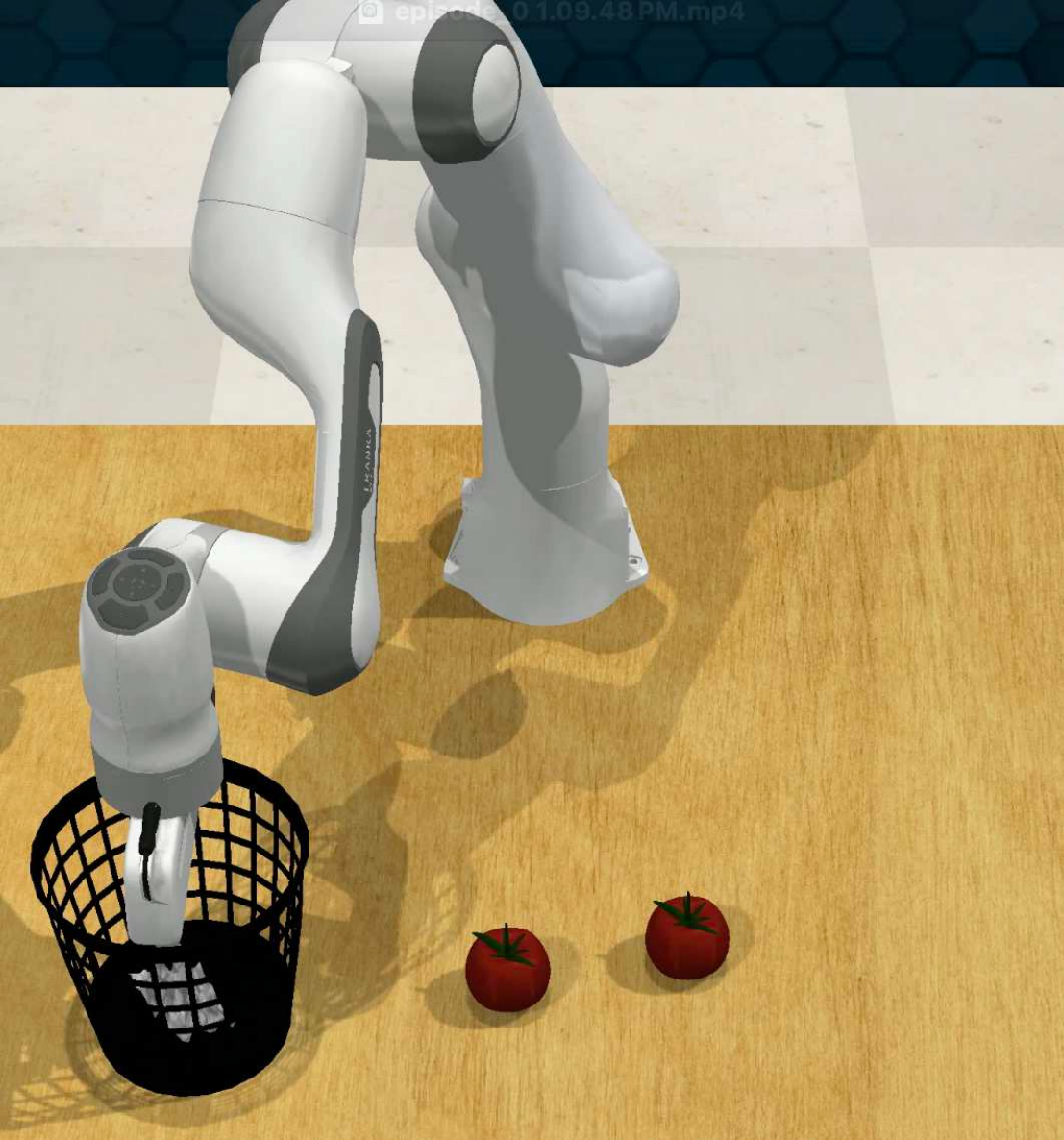}
      & \includegraphics[width=\linewidth,height=\linewidth,keepaspectratio=false]{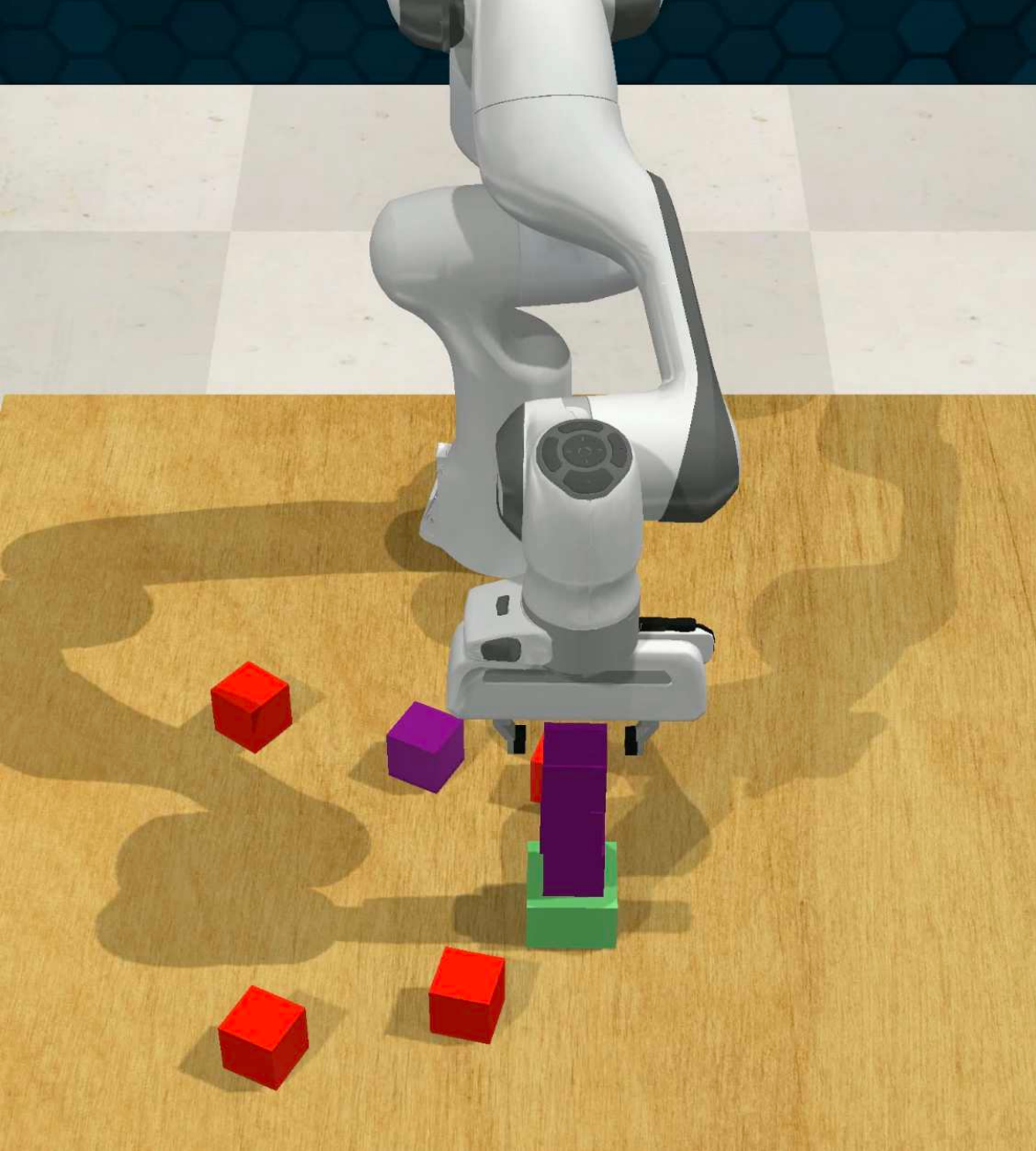} \\
\end{tabularx}
\caption{\textbf{Tasks in simulation.} We consider 9 pick-and-place tasks from RLBench simulator~\cite{James2019RLBenchTR}. For each task we display a starting condition (top) and the desired final state (bottom).}\label{fig:tasks-simulator}
\end{figure*}

\begin{table*}[t]
\setlength{\tabcolsep}{2pt}
\newcolumntype{C}{>{\centering\arraybackslash}p{1.2cm}}
\centering
\resizebox{\textwidth}{!}{
\begin{tabular}{l|CCCCCCCCC|C}
\toprule
\multirow{2}{*}{\textbf{Method}} 
& \textbf{Basketball} & \textbf{Close} & \textbf{Empty} & \textbf{Insert} & \textbf{Meat} & \textbf{Open} & \textbf{Put} & \textbf{Rubbish} & \textbf{Stack} & \multirow{2}{*}{\textbf{Average}} \\
& \textbf{in Hoop}   & \textbf{Jar}   & \textbf{Container} & \textbf{in Peg} & \textbf{off Grill} & \textbf{Bottle} & \textbf{Block} & \textbf{in Bin} & \textbf{Blocks} &  \\
\midrule
CAP~\cite{Liang2022CodeAP}       
& 0.0  & 0.0  & 0.0  & 8.0  & 0.0  & 0.0  & 76.0 & 0.0  & 0.0  & 9.3 \\

VoxPoser~\cite{Huang2023VoxPoserC3} 
& 20.0 & 0.0  & 0.0  & 0.0  & 0.0  & 0.0  & 36.0 & 64.0 & 32.0 & 16.9 \\
MALMM~\cite{Singh2024MALMMML}                            
& \underline{88.0} & 84.0 &\textbf{64.0} & 68.0 & \textbf{92.0} & \underline{96.0} & \textbf{100} & 80.0 & \underline{56.0} & \underline{80.9} \\
LLaMA-70B                        
& 70.0 & \textbf{99.0} & \underline{40.0} & \underline{86.0} & 66.0 & 93.0 & 93.0 & \textbf{97.0} & 49.0 & 77.0 \\
\midrule
LLaMA-8B                         
& 53.0 & 16.0 & 1.0  & 15.0 & 16.0 & 57.0 & 60.0 & 10.0 & 0.0  & 25.3 \\
$\Larrow$ w/ \framework          
    & \textbf{95.0} & \underline{88.0} & 39.0 & \textbf{97.0} & \underline{76.0} & \textbf{98.0} & \underline{98.0} & \underline{95.0} & \textbf{63.0} & \textbf{83.2} \\
\bottomrule
\end{tabular}}
\caption{\textbf{Comparison with zero-shot baselines.} We report the task success rate (\%) for different methods applied to the nine manipulation tasks from RLBench~\cite{James2019RLBenchTR}. With a small LLaMA-8B model finetuned with BLAZER, we are able to achieve the best performance. Note how LLaMA-8B with \framework outperforms considerably LLaMA-70B, that was used as $\text{LLM}_\text{boot}$. This implies that \framework can yields LLMs that outperform their teacher models on manipulation tasks. The table highlights the best-performing method for each task in \textbf{bold} and the second-best method is \underline{underlined}.}
\label{tab:task_comparison}
\end{table*}

\section{Experiments}
Our experiments are designed to assess the LLM agent trained using our \framework framework against existing vision and language foundation model-based baselines. 
\subsection{Experimental details}\label{sec:exp-baselines}
\noindent\textbf{Tasks and environment.} For comparison with baselines, we follow the MALMM setup and evaluate on nine simulated tasks from RLBench~\cite{James2019RLBenchTR}, illustrated in Fig.~\ref{fig:tasks-simulator}: \textit{Basketball in Hoop} (BH), \textit{Close Jar} (CJ), \textit{Empty Container} (EC), \textit{Insert in Peg} (IP), \textit{Meat off the Grill} (MG), \textit{Open Bottle} (OB), \textit{Put Block} (PB), \textit{Rubbish in Bin} (RB), and \textit{Stack Blocks} (SB). Detailed task descriptions and success conditions are provided in Appendix\ref{app:rlbench_tasks}. We train \framework by setting these tasks as $\mathcal{T}$. During testing, we randomize each test episode configuration, preventing overlaps with those used in training. We use CoppeliaSim and interface it with PyRep. In the real-world setup, we test generalization on 12 tasks on a tabletop using a 7-DOF Franka Emika Panda Research 3 robot equipped with a parallel jaw gripper. We use three Intel RealSense D435i RGB-D cameras to capture the frontal, right, and left views and the panda-py~\cite{Elsner2023TamingTP} library to control the robot arm.
\smallskip

\noindent\textbf{Implementation Details.} For our experiments, we choose LLaMA-3.1-8B (LLaMA-8B) as $\text{LLM}_\text{BLAZER}$, and LLaMA-3.3-70B (LLaMA-70B) as $\text{LLM}_\text{boot}$. The
prompt for data bootstrapping is adopted from Single Agent (SA) setup of MALMM~\cite{Singh2024MALMMML}, and is included in Appendix\ref{app:prompts}.
Both the LLMs, $\text{LLM}_\text{BLAZER}$ and $\text{LLM}_\text{boot}$ generate 3D waypoints for the gripper, while trajectories are
computed and executed using a motion planner, which is a common approach in RLBench~\cite{James2019RLBenchTR}. During SFT, the models are trained with a prompt completion loss, using $N=2 000$ examples per task. We train for 5 epochs with an effective batch size of 24. We adopted parameter-efficient finetuning via LoRA with a rank of 64 and a scaling factor ($\alpha$) of 16. The learning rate is 2e-5, with a cosine learning rate scheduler applied during training.
\smallskip


\noindent\textbf{Baselines.} We compare \framework to other methods using no manual supervision: CAP~\cite{Liang2022CodeAP}, VoxPoser~\cite{Huang2023VoxPoserC3} and MALMM~\cite{Singh2024MALMMML}. All baseline results are reported following MALMM~\cite{Singh2024MALMMML} and use GPT-4 Turbo~\cite{openai2023gpt4} as the underlying LLM model. Furthermore, we also test two zero-shot LLM baselines, which include LLaMA-70B and LLaMA-8B. To do so, we query each LLM with the prompt used for $\mathcal{D}_\text{BLAZER}$ generation, and directly apply the obtained $\mathcal{C}_\tau$ as policy. Note that in \framework we use LLaMA-70B as $\text{LLM}_\text{boot}$ and LLaMA-8B as $\text{LLM}_\text{BLAZER}$, so those baselines serve as comparison with respect to the base version of the LLMs used in our framework.

\begin{figure}[t]
  \centering
  \includegraphics[width=\linewidth]{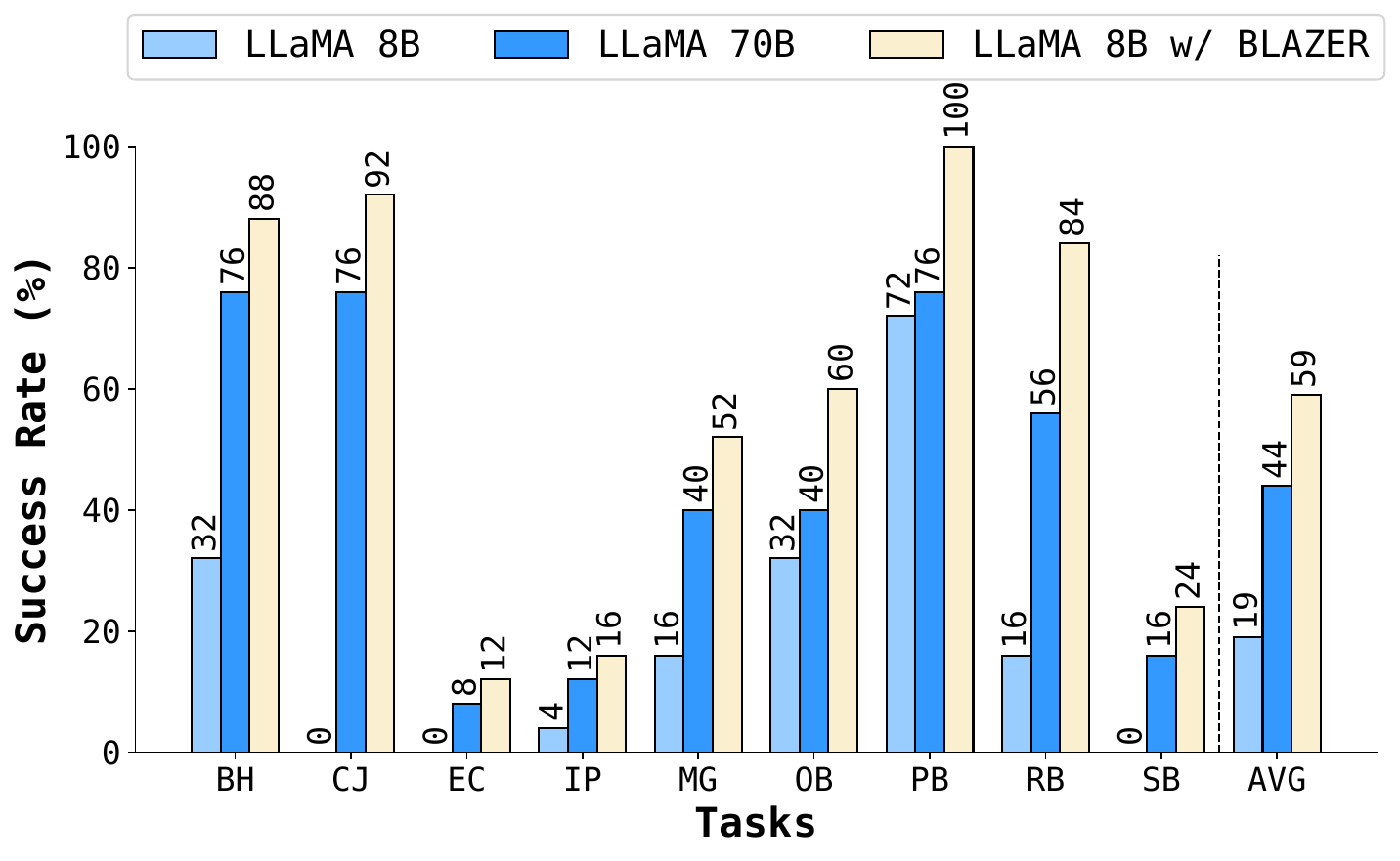}
\caption{\textbf{Tasks in simulation using visual observations.} We evaluate task success rate for LLaMA-70B, LLaMA-8B and LLaMA-8B w/ BLAZER in simulation using our vision pipeline that assuming no ground truth knowledge about object states. Consistently with results in Table~\ref{tab:task_comparison}, \framework outperform other methods.}
\label{fig:state_vs_vision}
\end{figure}

\subsection{Performance analysis}

\begin{figure*}[t]

\begin{subfigure}{\linewidth}
\centering
\setlength{\tabcolsep}{1pt}
\renewcommand{\arraystretch}{1}

\begin{tabularx}{\linewidth}{cY@{\hspace{3px}}Y@{\hspace{3px}}Y@{\hspace{7px}}Y@{\hspace{3px}}Y@{\hspace{3px}}Y@{\hspace{3px}}Y@{\hspace{3px}}Y@{\hspace{3px}}Y@{\hspace{3px}}}
&\multicolumn{3}{c}{\small{\textit{In-distribution tasks}}} 
& \multicolumn{6}{c}{\small{\textit{Out-of-distribution tasks}}}\\
\cmidrule(lr){2-4}\cmidrule(lr){5-10}

   & \shortstack{\footnotesize\textit{\mdseries Stack} \\      \footnotesize\textit{\mdseries Blocks}}
    & \shortstack{\footnotesize\textit{\mdseries Put} \\ \footnotesize\textit{\mdseries Block}}
    & \shortstack{\footnotesize\textit{\mdseries Rubbish} \\ \footnotesize\textit{\mdseries in Bin}}
    & \shortstack{\footnotesize\textit{\mdseries Block in} \\ \footnotesize\textit{\mdseries R/L Basket}}
    & \shortstack{\footnotesize\textit{\mdseries SW/SO Fruit} \\ \footnotesize\textit{\mdseries in Bowl}}
    & \shortstack{\footnotesize\textit{\mdseries Fruit in} \\ \footnotesize\textit{\mdseries Cl. Basket}}
    & \shortstack{\footnotesize\textit{\mdseries Cup on} \\ \footnotesize\textit{\mdseries C. Object}}
    & \shortstack{\footnotesize\textit{\mdseries Jar} \\ \footnotesize\textit{\mdseries in Bin}}
    & \shortstack{\footnotesize\textit{\mdseries Case on} \\ \footnotesize\textit{\mdseries Target}} \\
    \multirow{1}{*}[30pt]{\rotatebox{90}{\textit{Start}}}
      & \includegraphics[width=\linewidth,height=\linewidth,keepaspectratio=false]{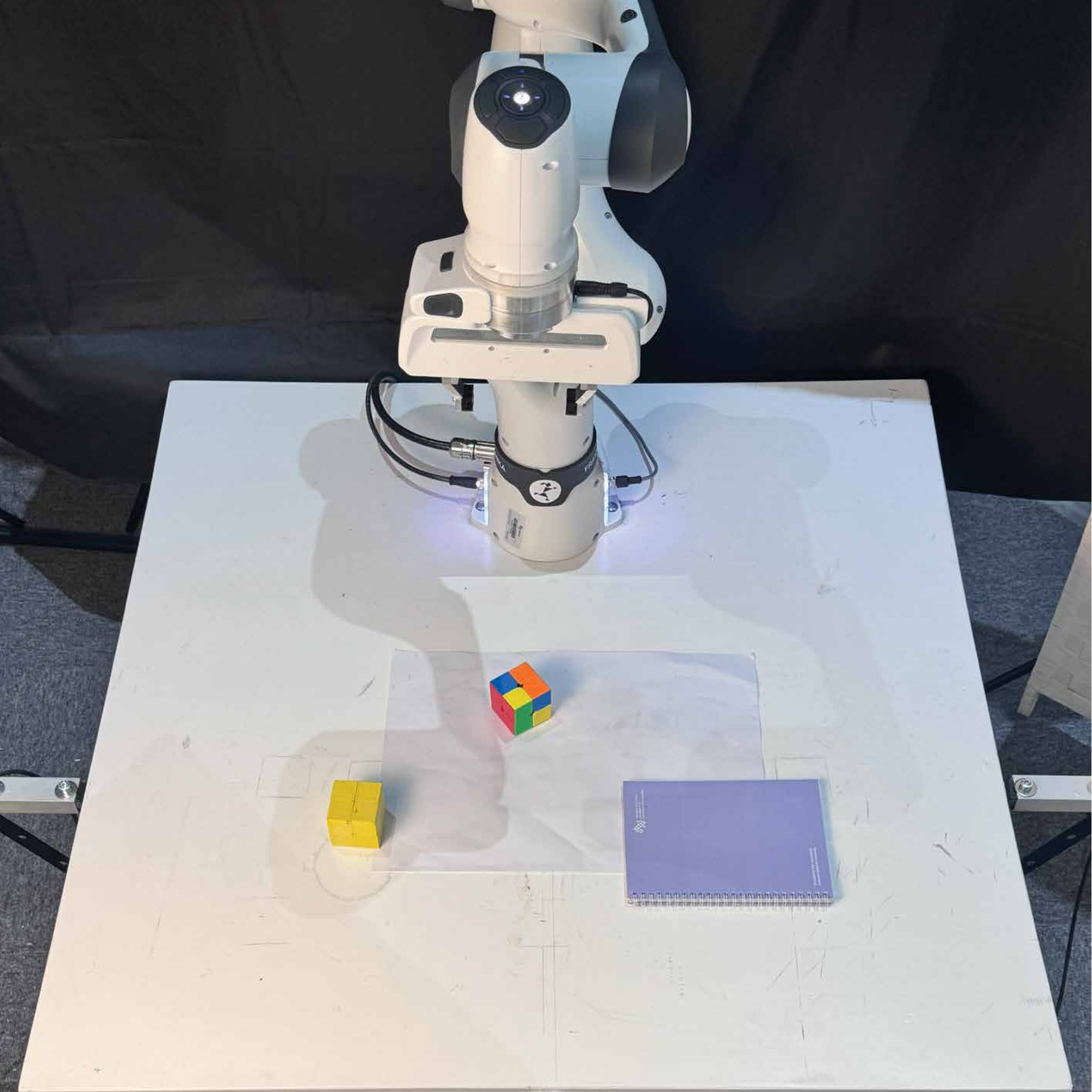}
      & \includegraphics[width=\linewidth,height=\linewidth,keepaspectratio=false]{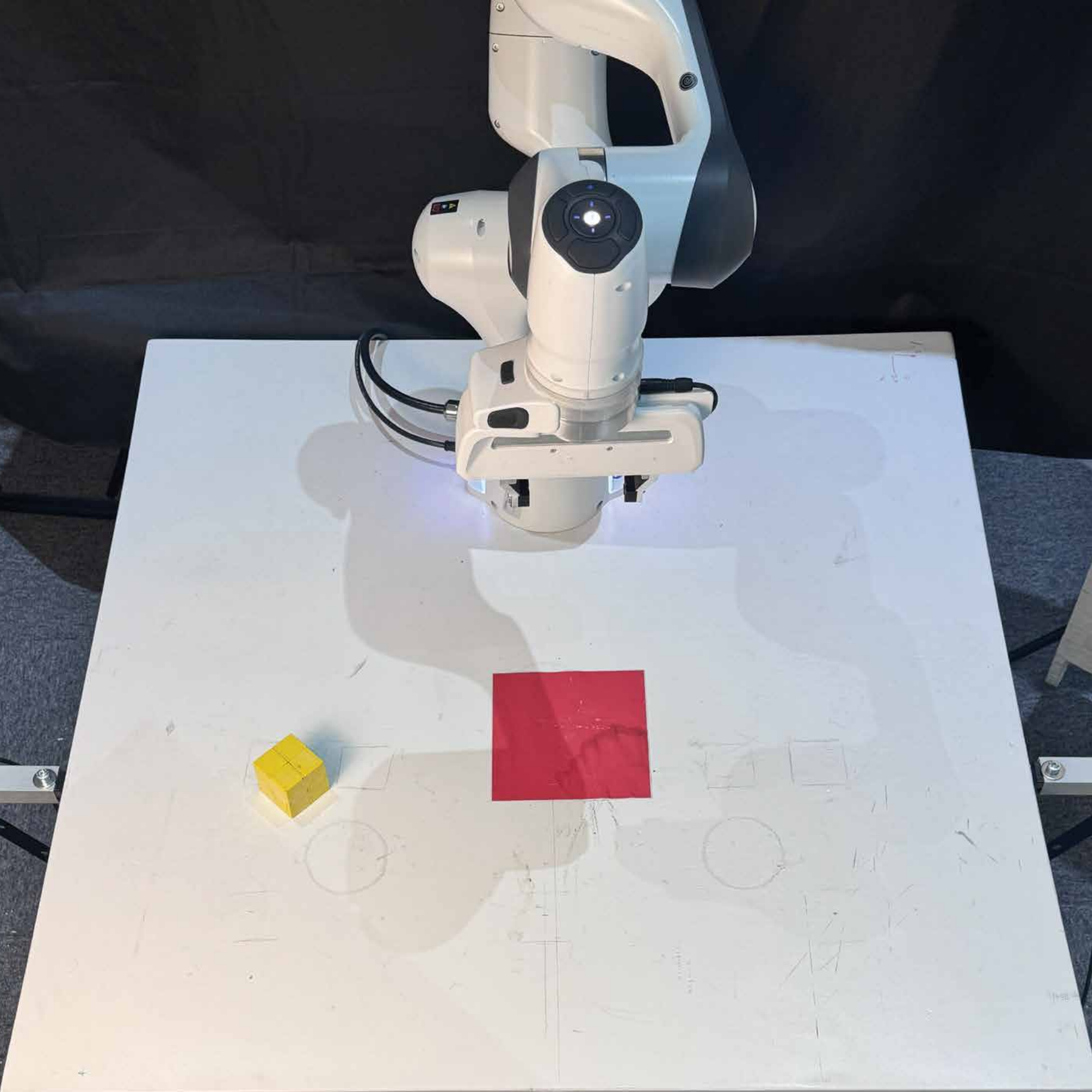}
      & \includegraphics[width=\linewidth,height=\linewidth,keepaspectratio=false]{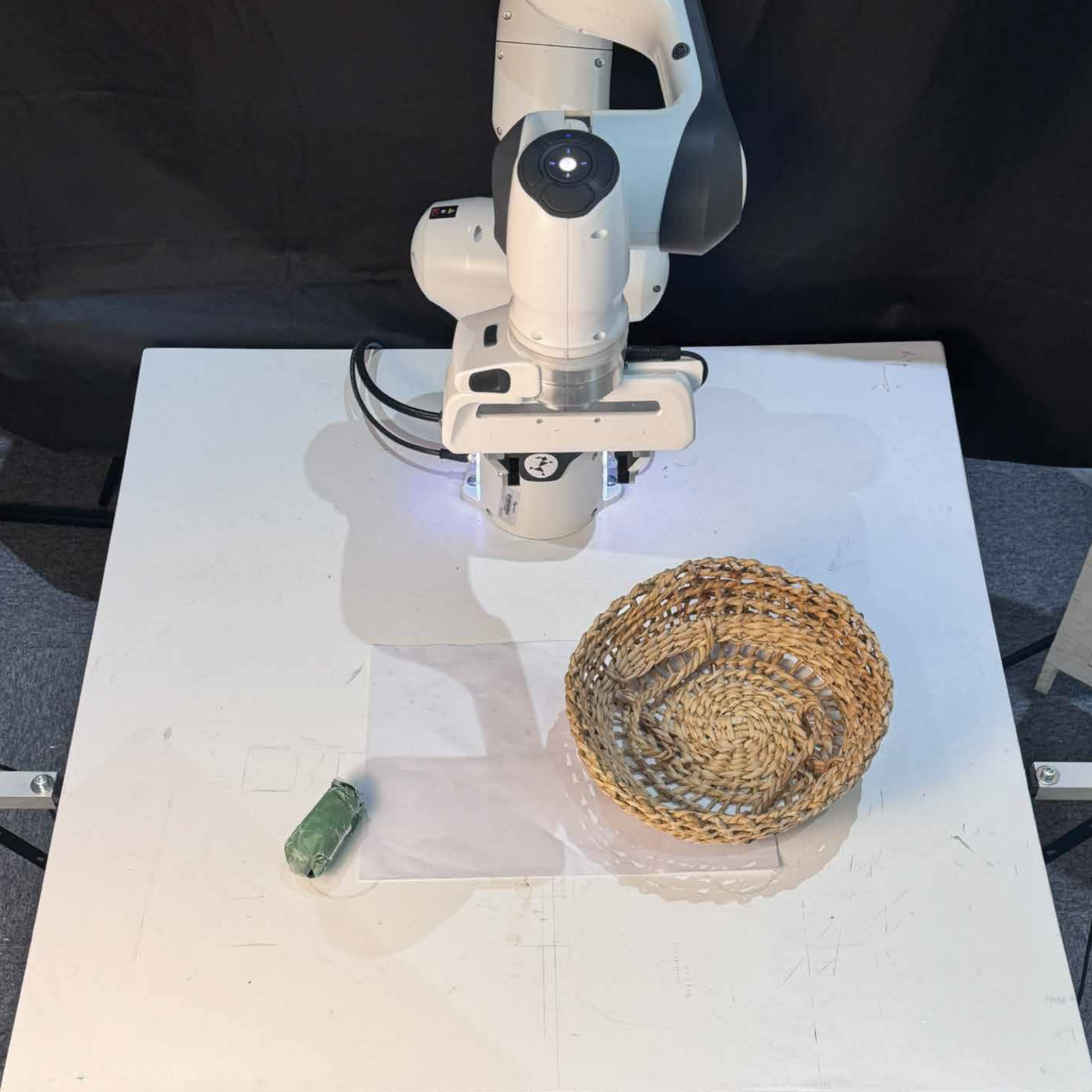}
      & \includegraphics[width=\linewidth,height=\linewidth,keepaspectratio=false]{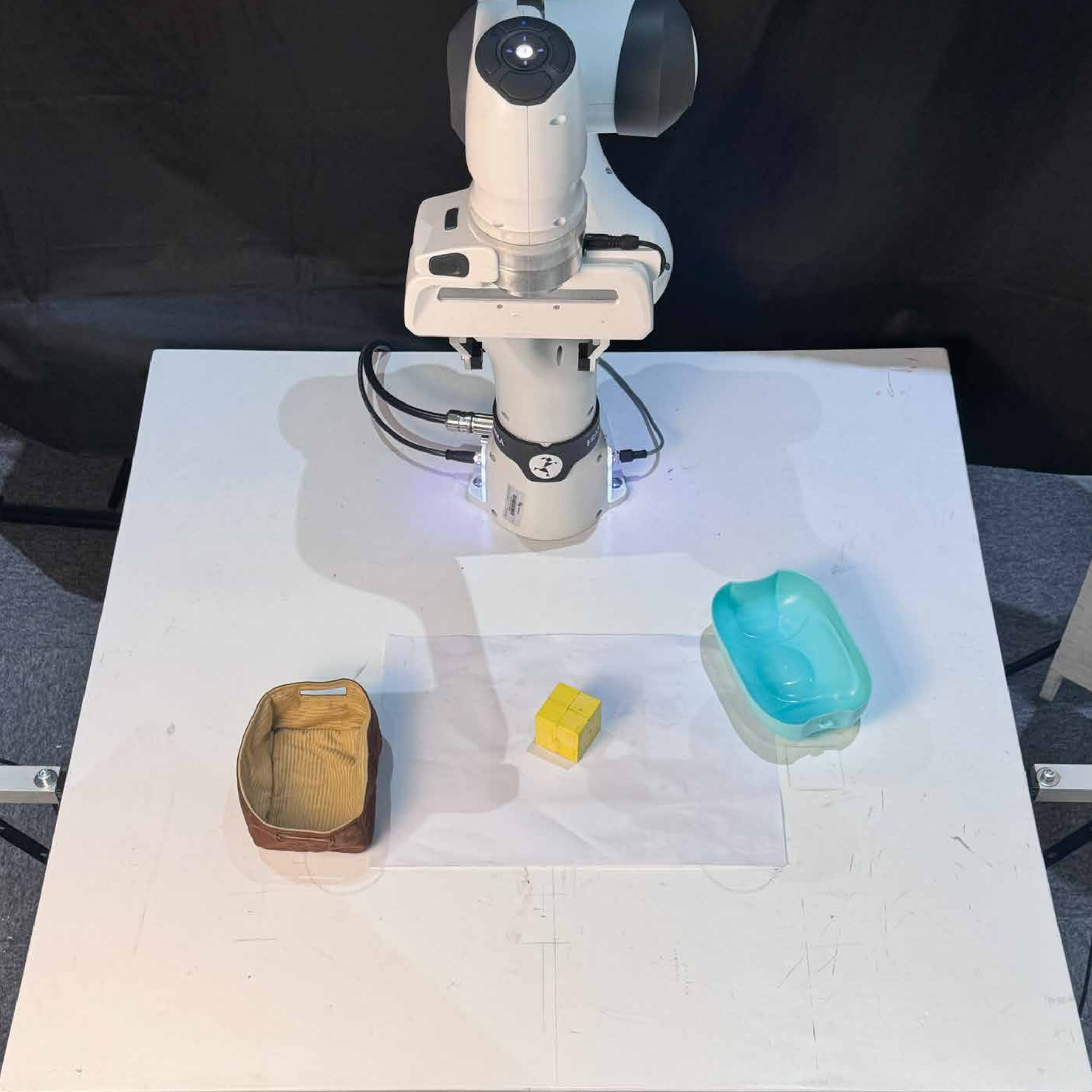}
      & \includegraphics[width=\linewidth,height=\linewidth,keepaspectratio=false]{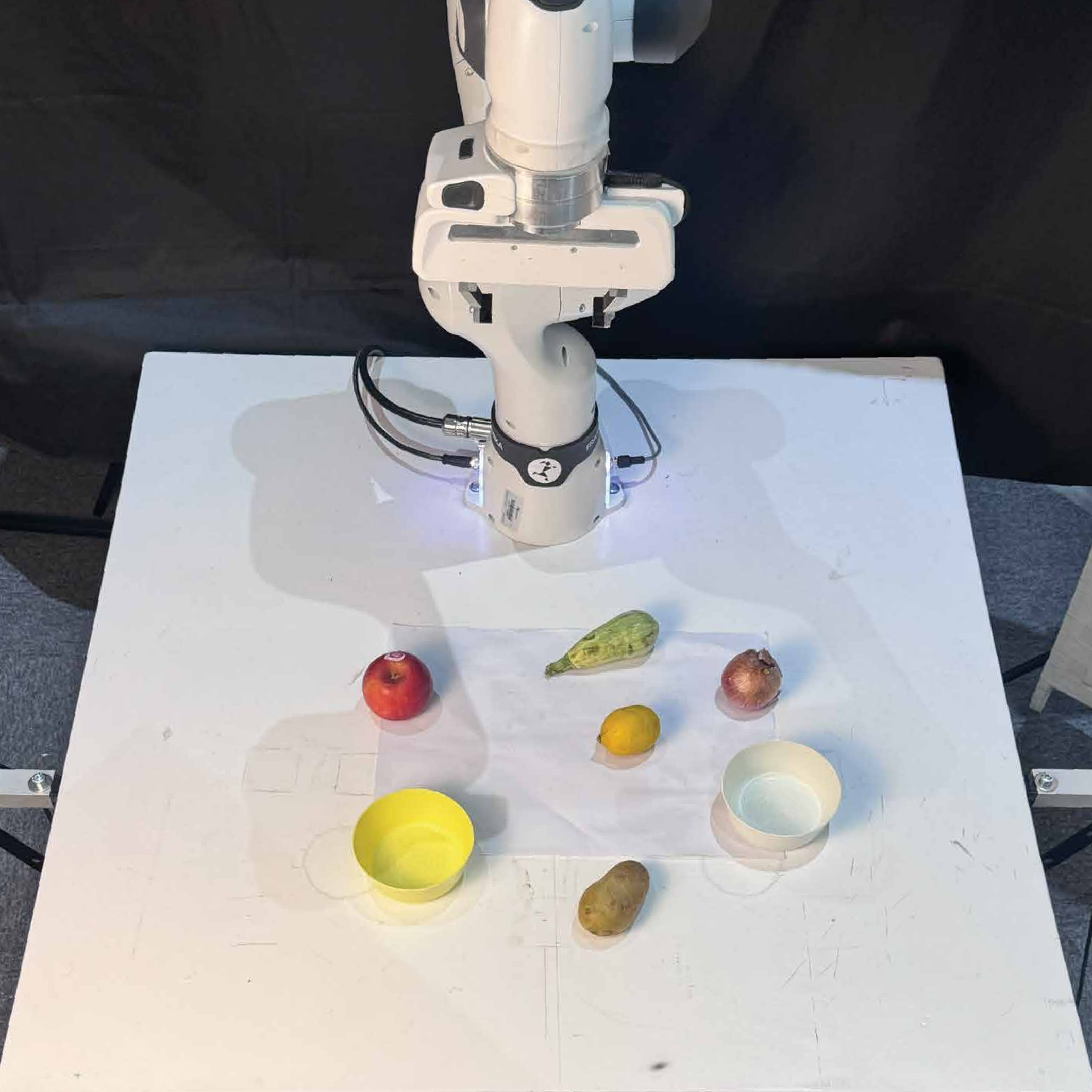}
      & \includegraphics[width=\linewidth,height=\linewidth,keepaspectratio=false]{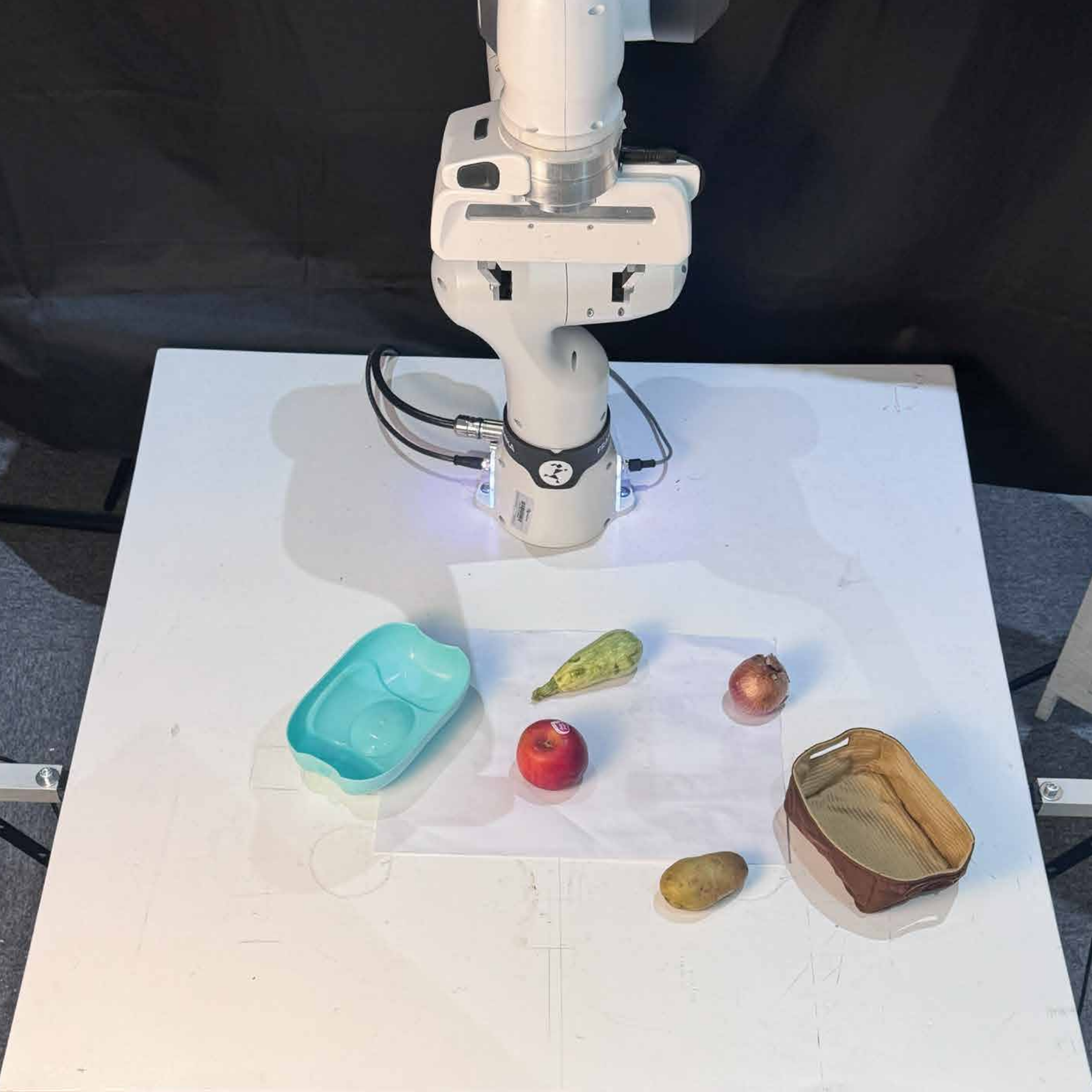}
      & \includegraphics[width=\linewidth,height=\linewidth,keepaspectratio=false]{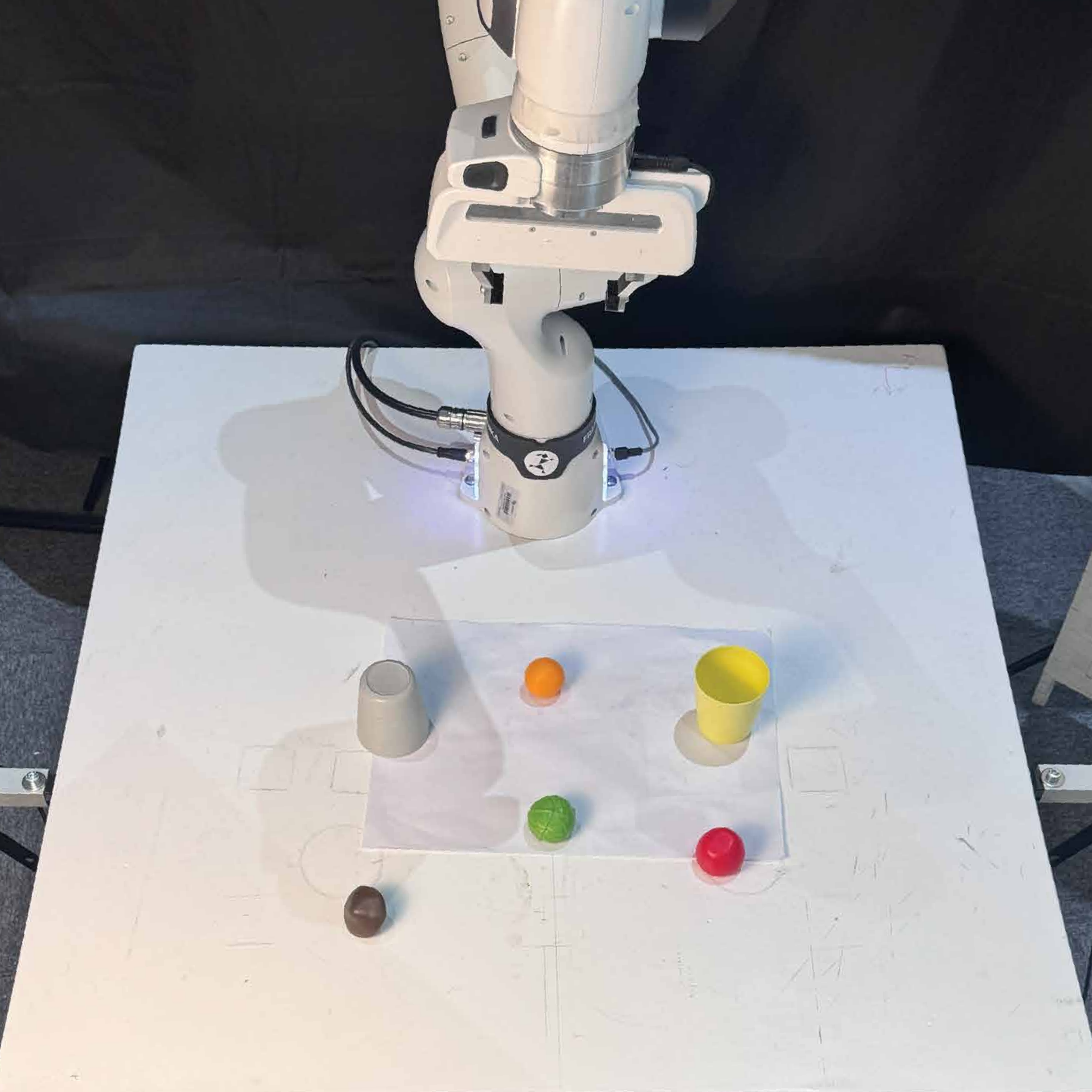}
      & \includegraphics[width=\linewidth,height=\linewidth,keepaspectratio=false]{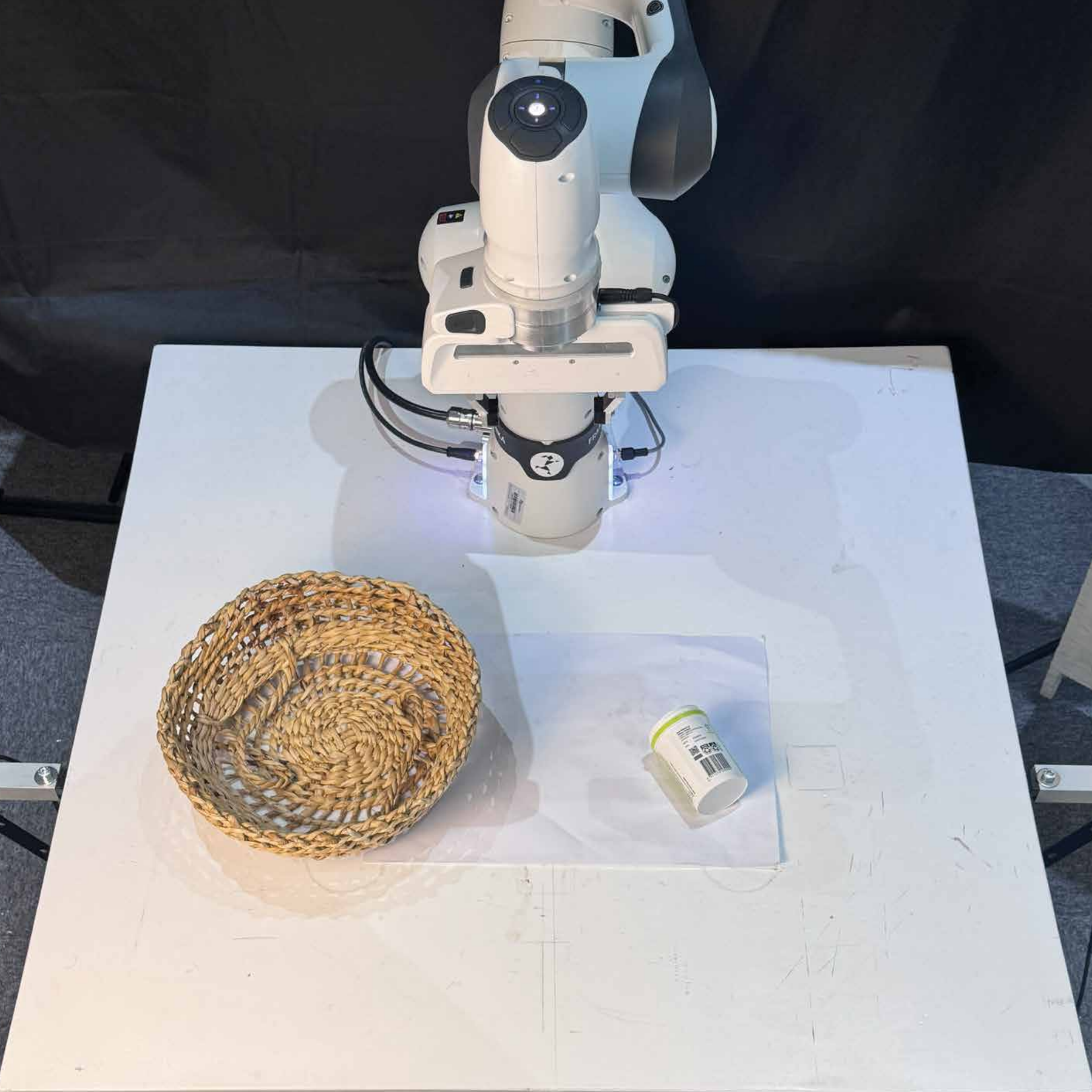}
      & \includegraphics[width=\linewidth,height=\linewidth,keepaspectratio=false]{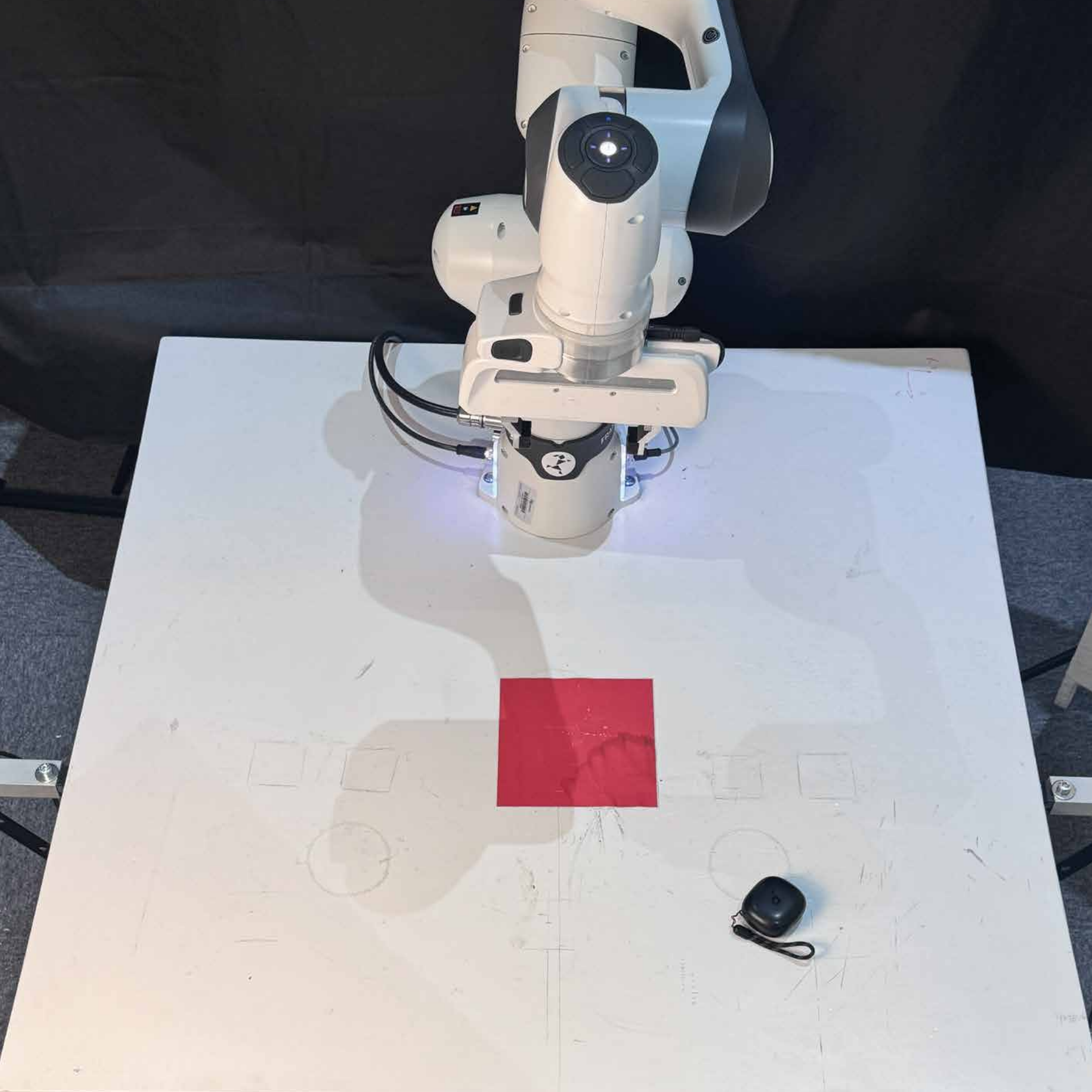} \\
    \multirow{1}{*}[28pt]{\rotatebox{90}{\textit{End}}}
      & \includegraphics[width=\linewidth,height=\linewidth,keepaspectratio=false]{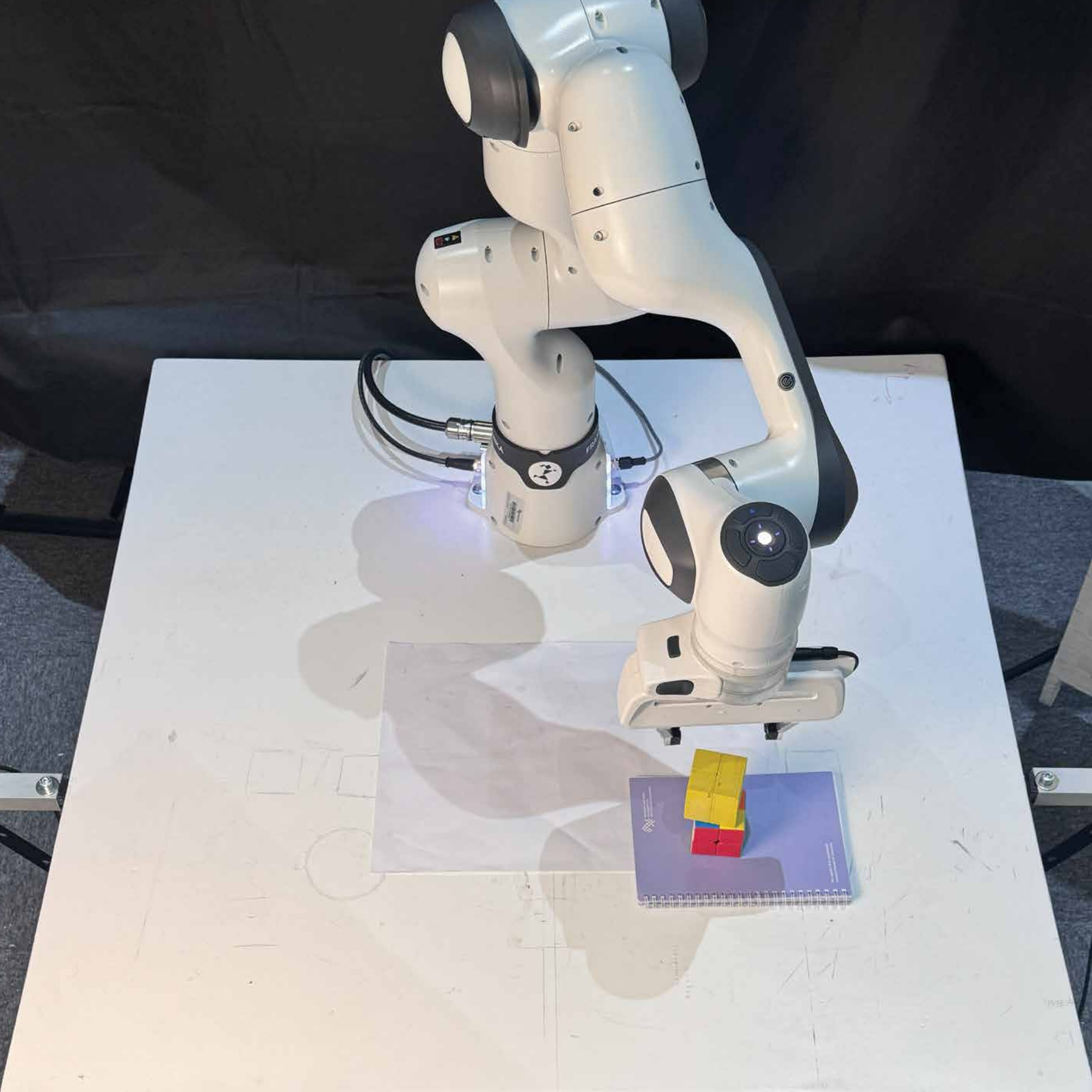}
      & \includegraphics[width=\linewidth,height=\linewidth,keepaspectratio=false]{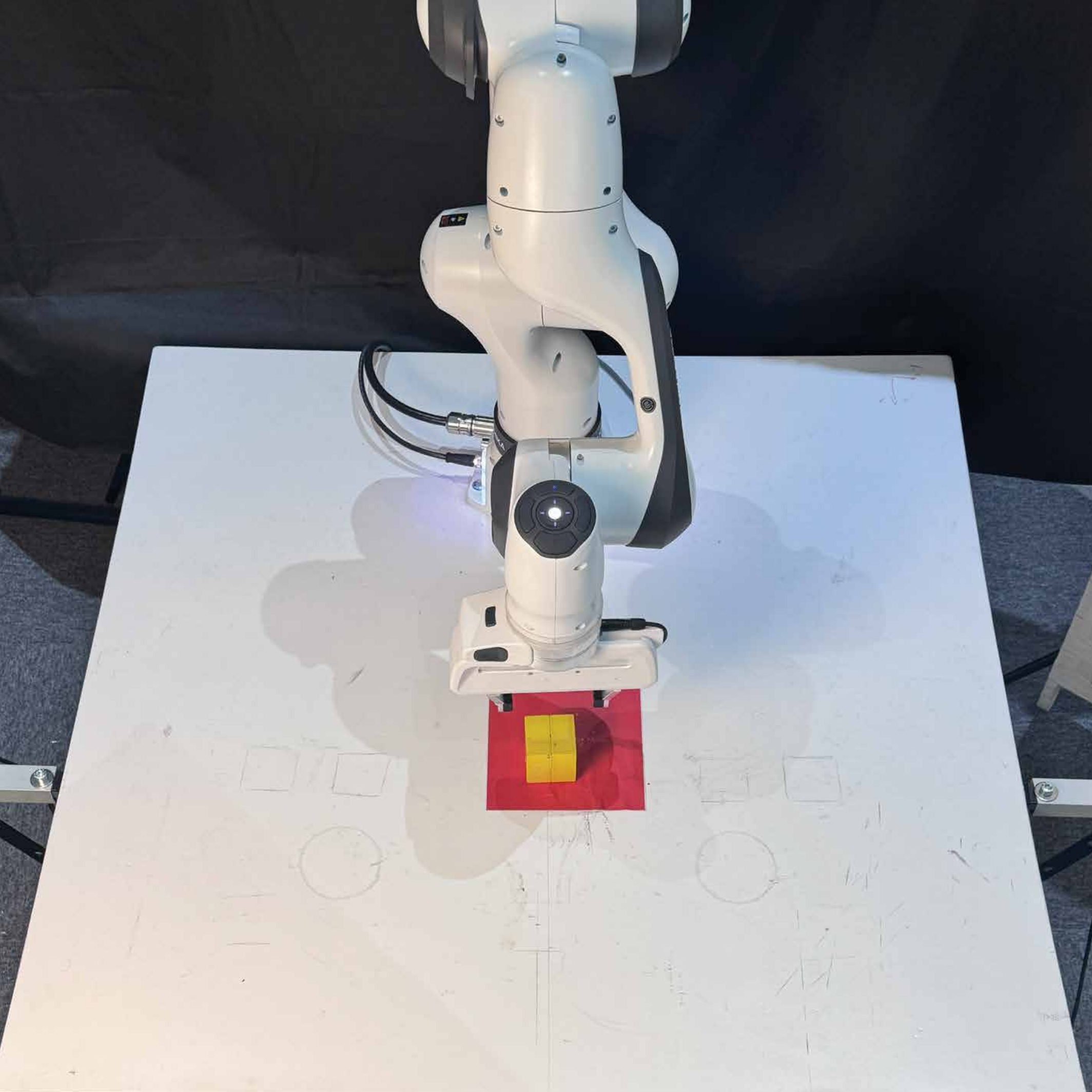}
      & \includegraphics[width=\linewidth,height=\linewidth,keepaspectratio=false]{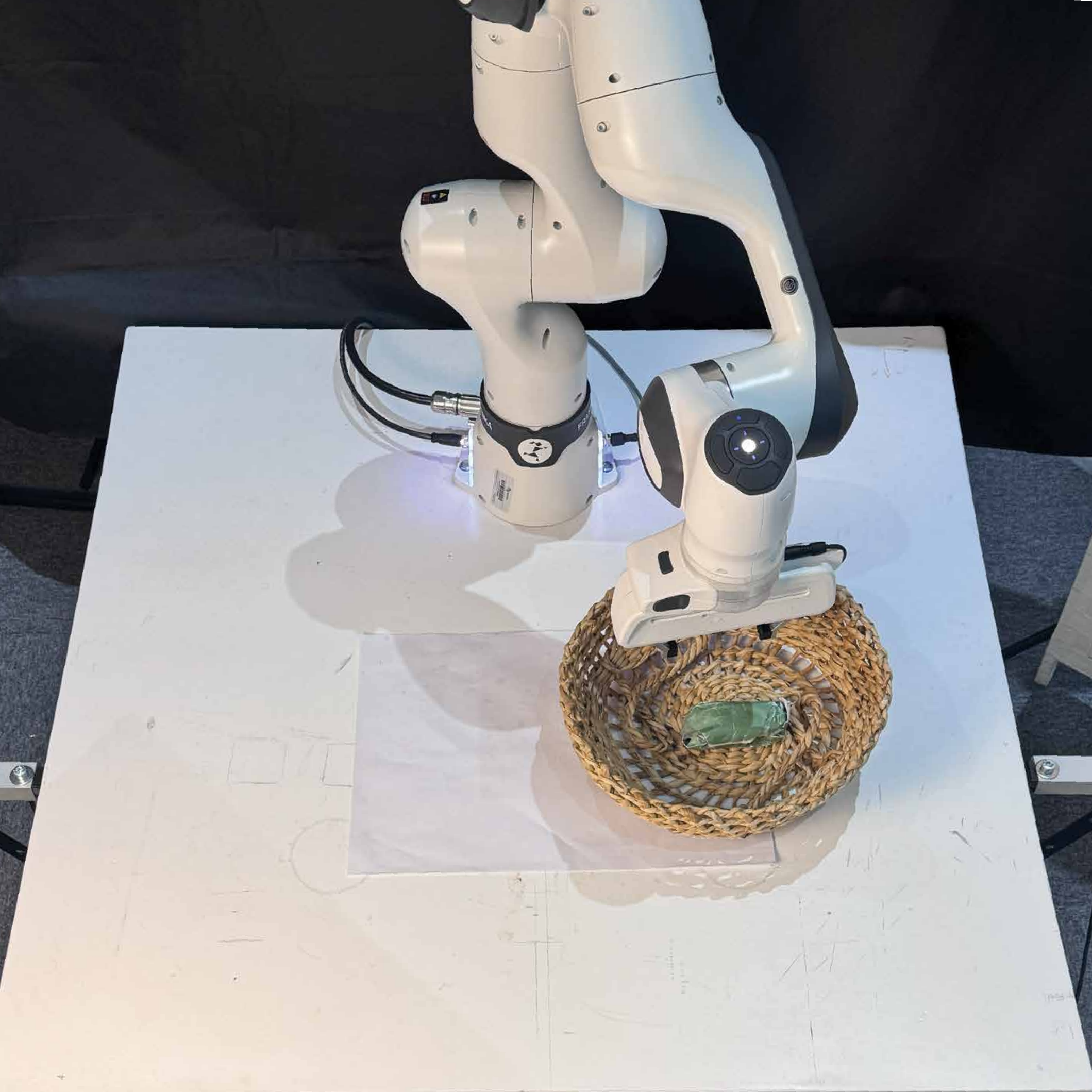}
      & \includegraphics[width=\linewidth,height=\linewidth,keepaspectratio=false]{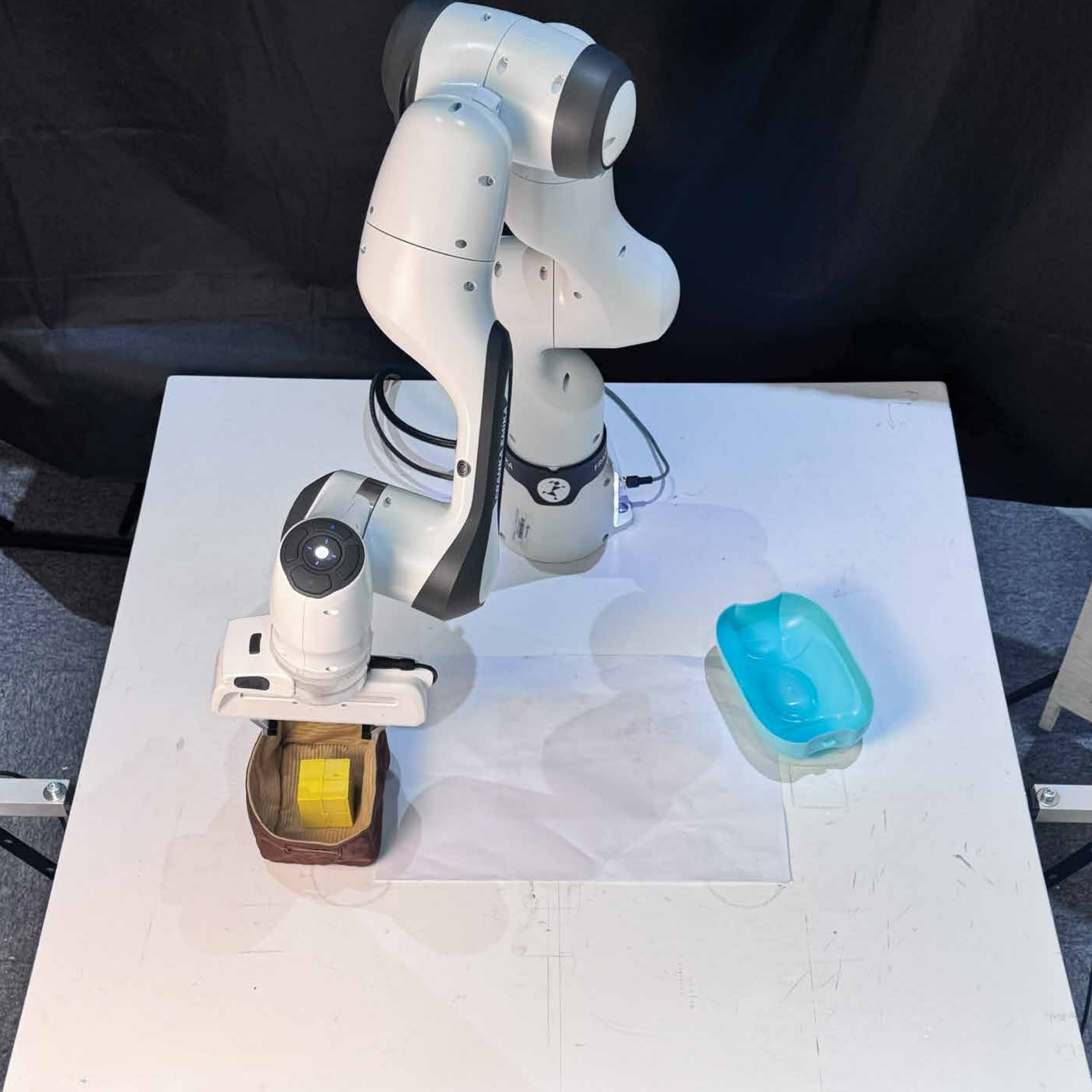}
      & \includegraphics[width=\linewidth,height=\linewidth,keepaspectratio=false]{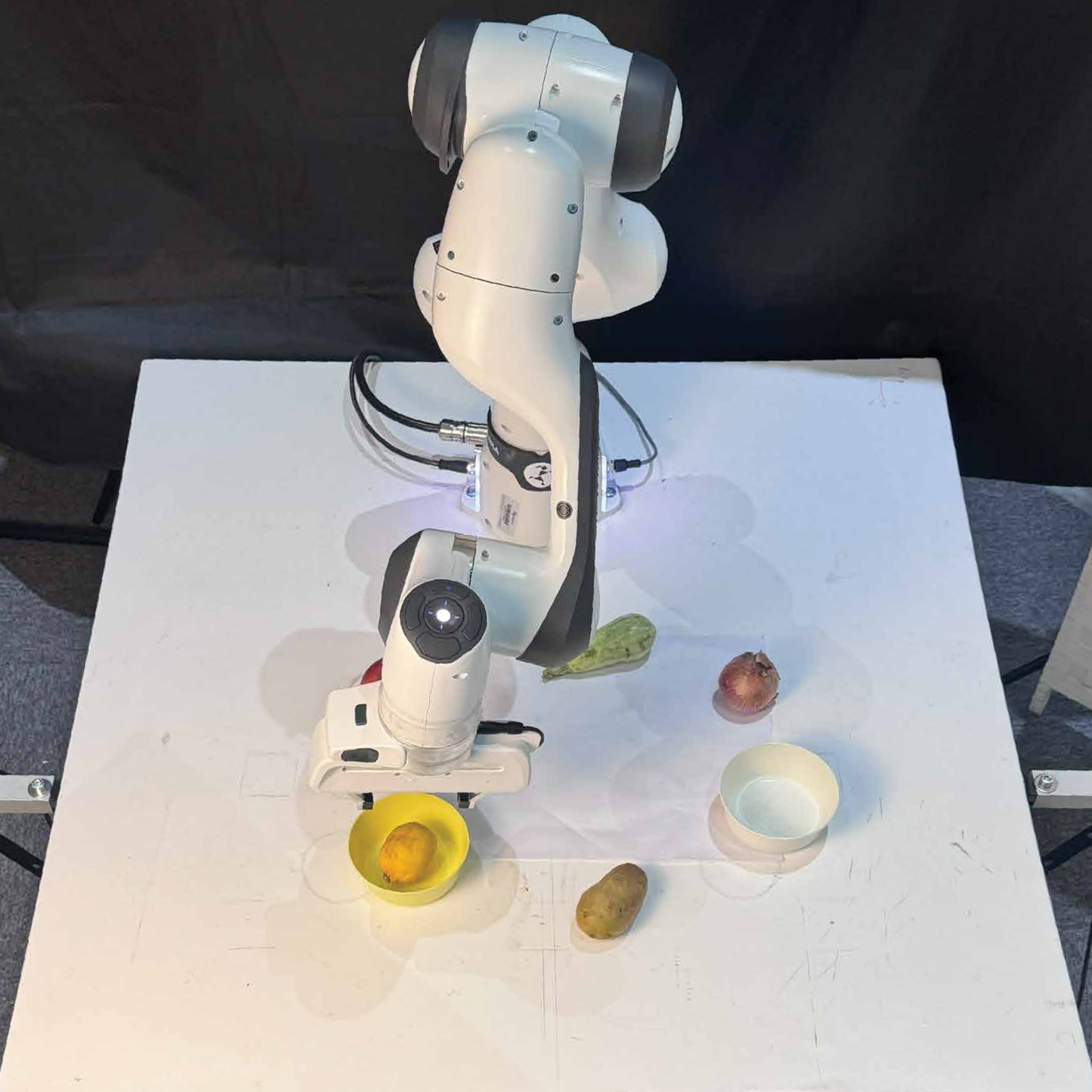}
      & \includegraphics[width=\linewidth,height=\linewidth,keepaspectratio=false]{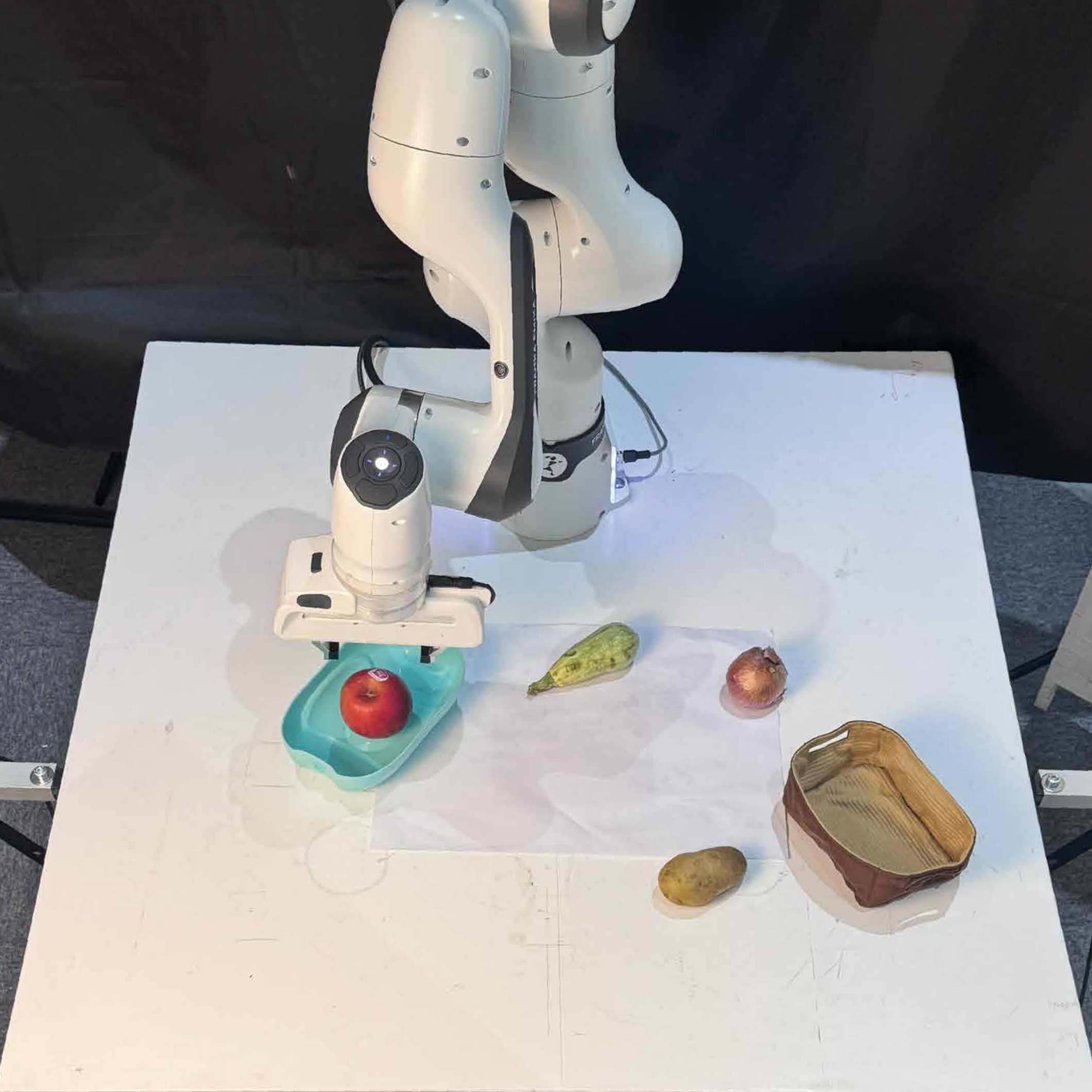}
      & \includegraphics[width=\linewidth,height=\linewidth,keepaspectratio=false]{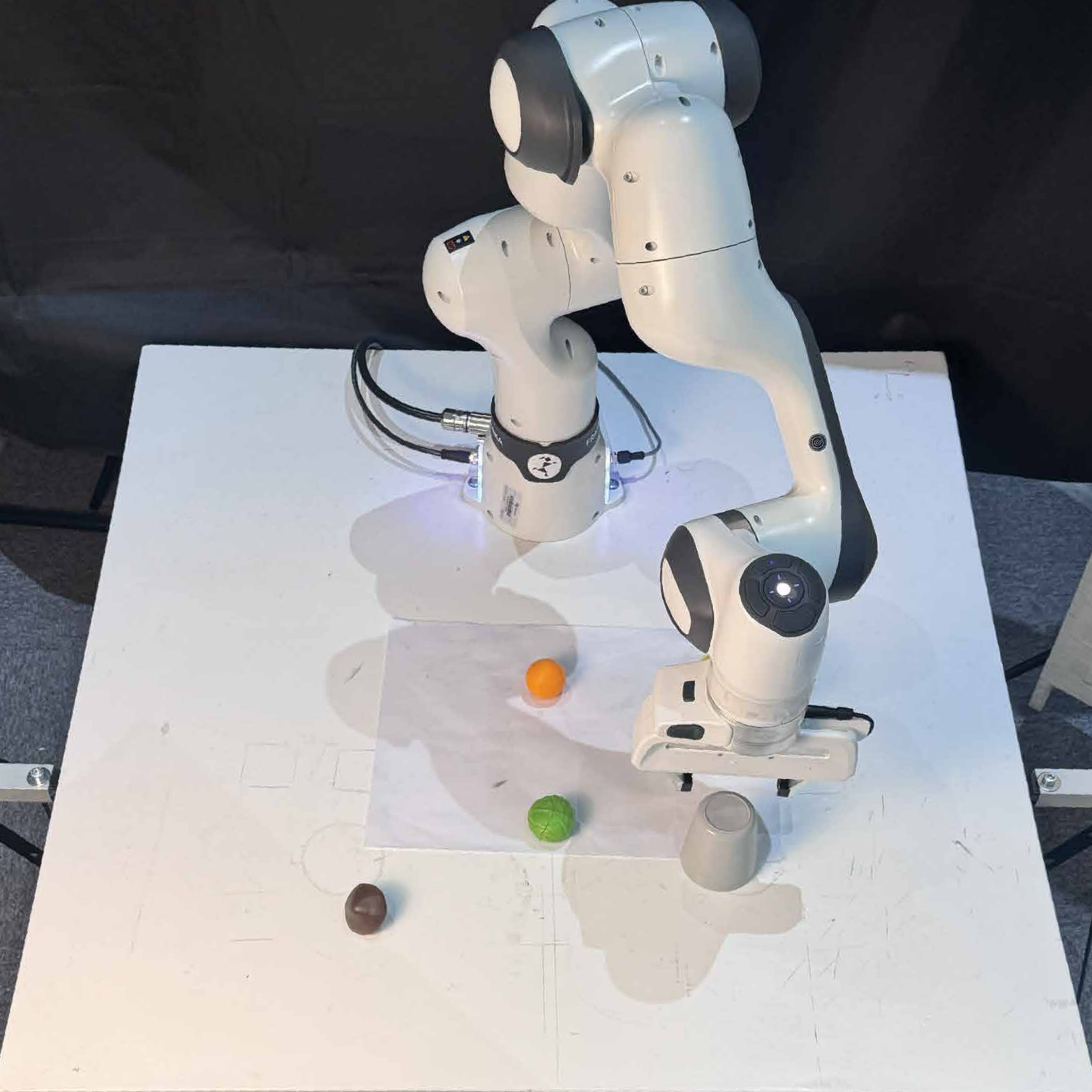}
      & \includegraphics[width=\linewidth,height=\linewidth,keepaspectratio=false]{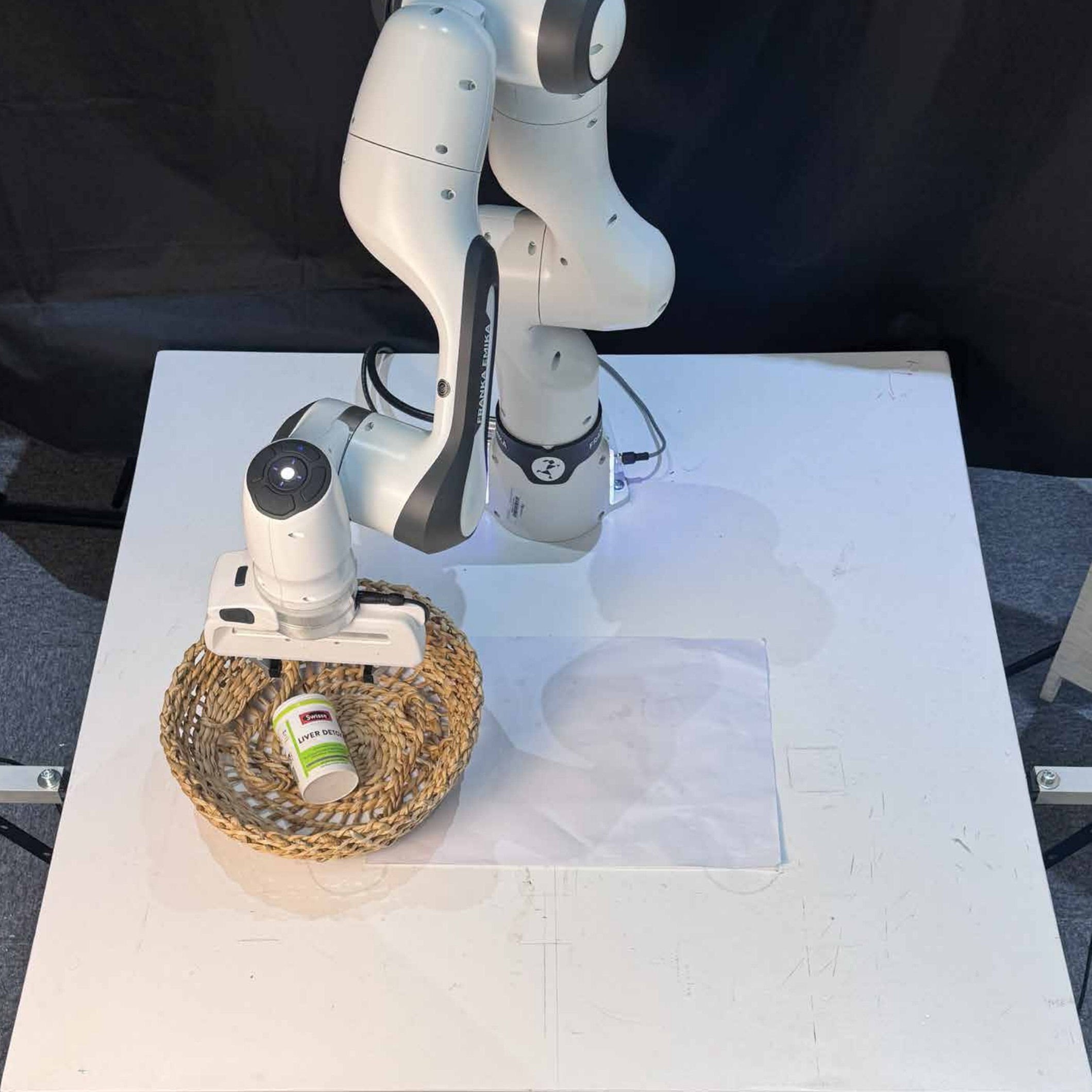}
      & \includegraphics[width=\linewidth,height=\linewidth,keepaspectratio=false]{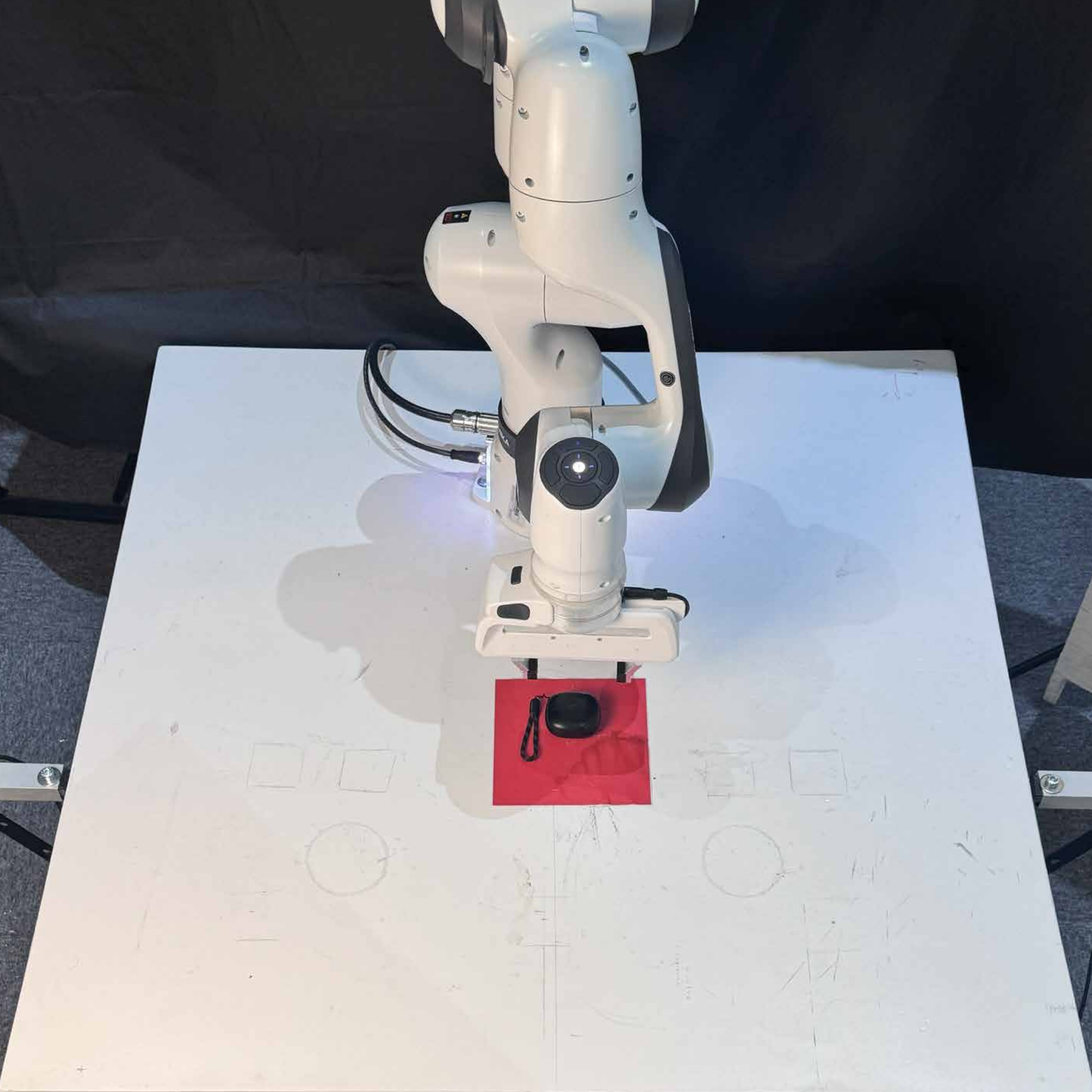} \\
\end{tabularx}
\caption{Real world tasks visualization}\label{fig:tasks-real-world-pic}
\end{subfigure}
\begin{subtable}{\linewidth}

\setlength{\tabcolsep}{4pt} 
\newcolumntype{C}{>{\centering\arraybackslash}p{1.6cm}} 
\centering
\resizebox{\textwidth}{!}{%
\begin{tabular}{l|CCC|CCCCCC|C}
\multicolumn{1}{c}{}& \multicolumn{3}{c}{\small\textit{In-distribution tasks}} 
& \multicolumn{6}{c}{\small\textit{Out-of-distribution tasks}} & \multicolumn{1}{c}{}\\
\toprule
& \small\bfseries Stack 
& \small\bfseries Put 
& \small\bfseries Rubbish 
& \small\bfseries Block in 
& \small\bfseries SW/SO 
& \small\bfseries Fruit in 
& \small\bfseries Cup on 
& \small\bfseries Jar 
& \small\bfseries Case 
& \small\bfseries \multirow{2}{*}{Average} \\
& \small\bfseries Blocks 
& \small\bfseries Block 
& \small\bfseries in Bin 
& \small\bfseries R/L Basket 
& \small\bfseries Fruit in B. 
& \small\bfseries Cl. Basket 
& \small\bfseries C. Object 
& \small\bfseries in Bin 
& \small\bfseries on Target 
& \\
\midrule
LLaMA-70B 
& 0.3 {\scriptsize\textcolor{gray}{(3/10)}} 
& 0.4 {\scriptsize\textcolor{gray}{(4/10)}} 
& 0.2 {\scriptsize\textcolor{gray}{(2/10)}} 
& \textbf{0.6 {\scriptsize\textcolor{gray}{(6/10)}}} 
& \textbf{0.4 {\scriptsize\textcolor{gray}{(4/10)}}} 
& 0.5 {\scriptsize\textcolor{gray}{(5/10)}} 
& 0.3 {\scriptsize\textcolor{gray}{(3/10)}} 
& 0.3 {\scriptsize\textcolor{gray}{(3/10)}} 
& 0.1 {\scriptsize\textcolor{gray}{(1/10)}} 
& 0.333 \\
LLaMA-8B w/ \textsc{BLAZER} 
& \textbf{0.4 {\scriptsize\textcolor{gray}{(4/10)}}} 
& \textbf{0.6 {\scriptsize\textcolor{gray}{(6/10)}}} 
& \textbf{0.4 {\scriptsize\textcolor{gray}{(4/10)}}} 
& 0.5 {\scriptsize\textcolor{gray}{(5/10)}} 
& \textbf{0.4 {\scriptsize\textcolor{gray}{(4/10)}}} 
& \textbf{0.7 {\scriptsize\textcolor{gray}{(7/10)}}} 
& \textbf{0.5 {\scriptsize\textcolor{gray}{(5/10)}}} 
& \textbf{0.4 {\scriptsize\textcolor{gray}{(4/10)}}} 
& \textbf{0.4 {\scriptsize\textcolor{gray}{(4/10)}}} 
& \textbf{0.478} \\
\bottomrule
\end{tabular}
}
\caption{Quantitative evaluation.}
\label{tab:tasks-real-world-quant}
\end{subtable}
\caption{\textbf{Real world results.} We compare LLaMA-8B with \framework against LLaMA-70B on real-world tasks depicted in (\subref{fig:tasks-real-world-pic}). From quantitative results in (\subref{tab:tasks-real-world-quant}), we outperform the baseline, both on In-distribution tasks (similar to $\mathcal{T}$) and Out-of-distribution tasks, showcasing the generalization capability of \framework.}
\end{figure*}

\noindent\textbf{Comparison to baselines}
In Table~\ref{tab:task_comparison}, we present the performance of our $\text{LLM}_\text{\framework}$ model on 100 episodes for each task in $\mathcal{T}$, along with results of baselines introduced in Sec.~\ref{sec:exp-baselines}. Here we test all models assuming the knowledge of ground truth states $\Sigma_\mathcal{E}$ provided by the simulator.

Our LLaMA-8B model trained with \framework \textit{outperforms all baselines} on average (\textbf{83.2\%} success). This shows that the bootstrapping of reasoning data for manipulation tasks and the training of LLMs leads to stronger agents for robotic manipulation, proving the effectiveness of \framework. Importantly, our LLaMA-8B model with \framework surpasses considerably (\textbf{+6.2\%} on average) LLaMA-70B, which we used as $\text{LLM}_\text{boot}$, using only a fraction of the parameters. Moreover, it substantially increases the capabilities of the base LLaMA-8B (\textbf{+58.4\%}). Even in long-term tasks such as \textit{Stack Blocks} and \textit{Empty Container}, LLaMA-8B trained with \framework outperforms LLaMA-70B in \textit{Stack Blocks} by 14\% and is almost on par with it on the \textit{Empty Container} task. We notice that LLaMA-70B shows remarkable results with our detailed prompt, outperforming Code-as-Policy and VoxPoser, and proving the strength of our baseline. The MALMM multi-agent framework still surpasses LLaMA-70B due to its multistep failure detection and correction approach, although it still falls short of our agent.

Interestingly, we also note that zero-shot usage of LLaMA-8B results in unsatisfactory performance (avg. \textbf{25.3\%} success). In particular, for \textit{Stack Blocks} and \textit{Empty Container}, LLaMA-8B often fails to generate valid control commands, even for a single episode in \textit{Stack Blocks}. In contrast, LLaMA-70B successfully handles most tasks. This result is consistent with recent studies on long-context generation in large language models~\cite{An2024WhyDT, Hsieh2024RULERWT}. Ultimately, the ability of LLaMA-70B to provide more successful examples for $\mathcal{D}_\text{\framework}$ justifies its usage as $\text{LLM}_\text{boot}$.
\smallskip

\noindent\textbf{Perception impact.} We now evaluate the robustness of \framework to a more realistic setup with visual observations replacing ground truth object states. We employ our vision-based pipeline (Section~\ref{sec:method-perception}) to process input views for simulated tasks in Table~\ref{tab:task_comparison}, and obtain $\tilde{\Sigma}_\mathcal{E}$, which we use as input for LLMs. For fair comparison, we equip LLaMA-70B and the base LLaMA-8B as well as \framework with the same vision pipeline and report results in Figure~\ref{fig:state_vs_vision}. We observe that even with the noisy state estimation of our vision pipeline, LLaMA-8B trained with \framework still outperforms alternatives. In particular, the gain with respect to LLaMA-70B (\textbf{+15\%}) is even higher than when providing the ground truth $\Sigma_\mathcal{E}$ (\textbf{+6.2\%}, see Table~\ref{tab:task_comparison}). This shows the robustness to noise in the policies generated with LLMs trained with \framework.\looseness=-1

\subsection{Robot experiments}
We next deploy \framework in the real robot setup and compare its performance to LLaMA-70B baseline while using the same vision pipeline for both methods.
\smallskip

\begin{figure*}[ht]
 \centering
 \includegraphics[trim={0.5cm 1.5cm 2.7cm 0cm},width=1\linewidth]
 {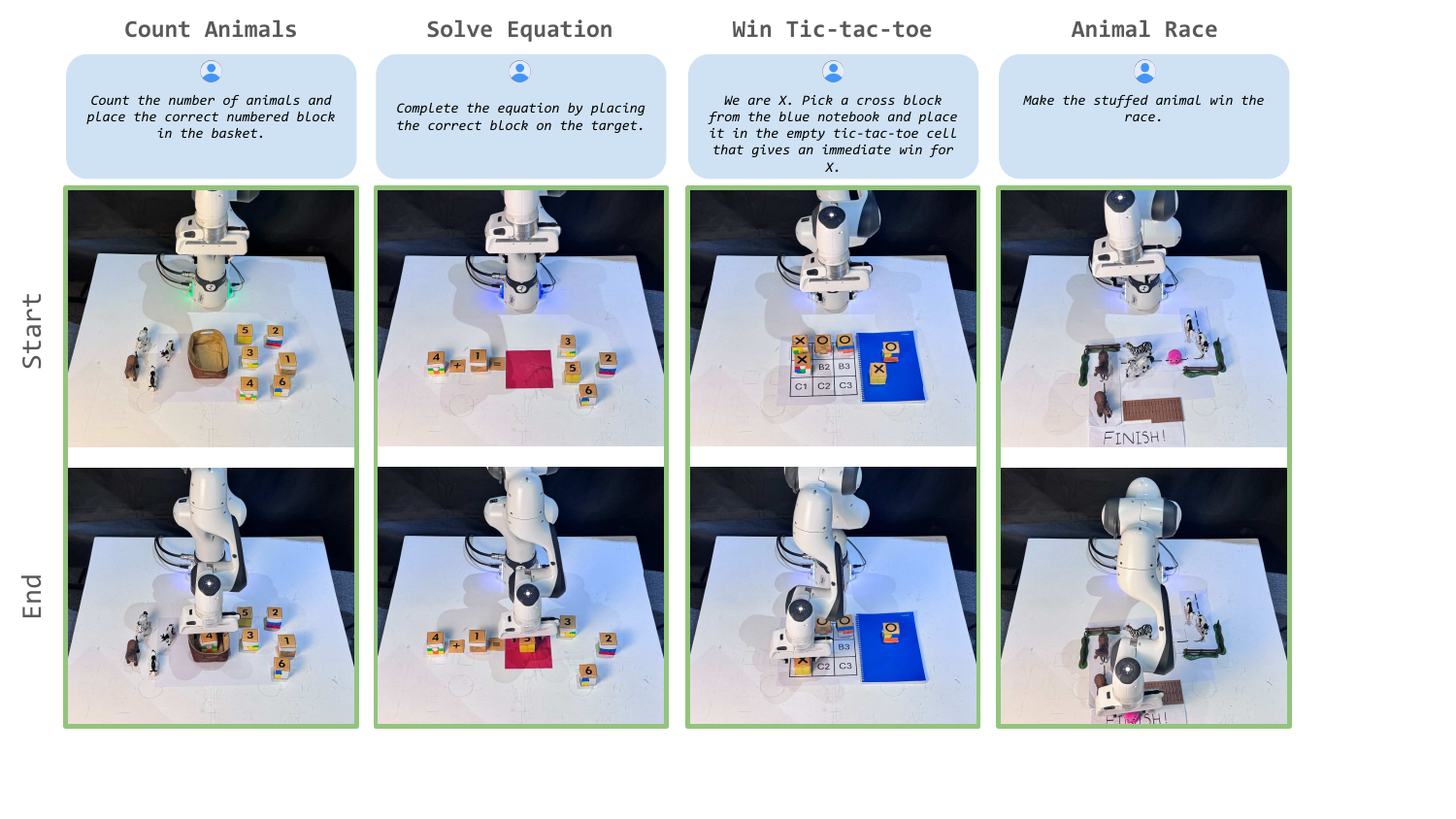}
 \caption{\textbf{Real-world results on reasoning tasks.} We show examples of four tasks that illustrate high-level reasoning and planning capabilities of BLAZER and demonstrate its generalization to new tasks. We also provide the prompt to \framework in the blue box.}
 \label{fig:tasks-real-world}
\end{figure*}

\noindent\textbf{Generalization capabilities.}
We use 9 additional tasks, shown in Figure~\ref{fig:tasks-real-world-pic}, to assess the transferability of \framework to real manipulation scenarios. We first consider \textit{In-distribution} tasks that resemble the \textit{Stack Blocks}, \textit{Put Block}, and \textit{Rubbish in Bin} tasks in Figure~\ref{fig:tasks-simulator}. We use these tasks to quantify the transfer of tasks in $\mathcal{T}$ in real-world deployment. We also propose 6 new \textit{Out-of-distribution} tasks, different from those in $\mathcal{T}$: \textit{Block in Right/Left Basket}, \textit{Sweet/Sour Fruit in Bowl}, \textit{Fruit in Closest Basket}, \textit{Cup on Colored Object}, \textit{Jar in Bin}, and \textit{Case on Target}. By testing these tasks, we aim to demonstrate the ability of \framework to solve tasks \textit{beyond} those in $\mathcal{T}$. We average results over 10 episodes for each task, replacing generic attributes in the task description (e.g. ``colored'', ''sweet/sour'') with precise values (e.g. ``red'', ``sour''), randomized for each episode. From the results in Table~\ref{tab:tasks-real-world-quant}, we show that \textit{our LLM agent trained with \framework still outperforms the baseline}, proving that the capabilities of tasks in $\mathcal{T}$ transfer successfully to the real world. In particular, for \textit{in-distribution} tasks, we notice a higher \textit{Stack Blocks} performance, that was suboptimal in Figure~\ref{fig:state_vs_vision}. We speculate that the performance in simulation could have been influenced mainly by the presence of distractor elements (see Figure~\ref{fig:tasks-simulator}). For out-of-distribution tasks, we beat LLaMA-70B on 5 tasks out of 6, ultimately proving that the benefits  of \framework training transfer to new task in real world settings.


\vspace{4px}\noindent\textbf{Qualitative examples.} 
To further evaluate the generalization abilities of \framework, we present it with four tasks that require high-level reasoning.
As in previous experiments in Fig.~\ref{tab:tasks-real-world-quant}, we use \framework trained in simulation on $\mathcal{T}$ tasks together with the vision pipeline described in Section~\ref{sec:method-perception}.
Specifically, we employ numbered blocks 
to test math capabilities 
in the \textit{Count Animals} and \textit{Solve Equation} tasks, game strategic planning in \textit{Win Tic-tac-toe}, and contextual awareness in \textit{Animal Race}. We provide qualitative results of \framework for these tasks in Figure~\ref{fig:tasks-real-world}. As can be seen, \framework can successfully solve tasks that require high-level reasoning while being substantially different from the training tasks $\mathcal{T}$. Videos illustrating execution of these tasks can be found on the project website~\cite{blazerwebpage}.\looseness=-1


\begin{figure}[t]
\centering
\includegraphics[width=\linewidth]{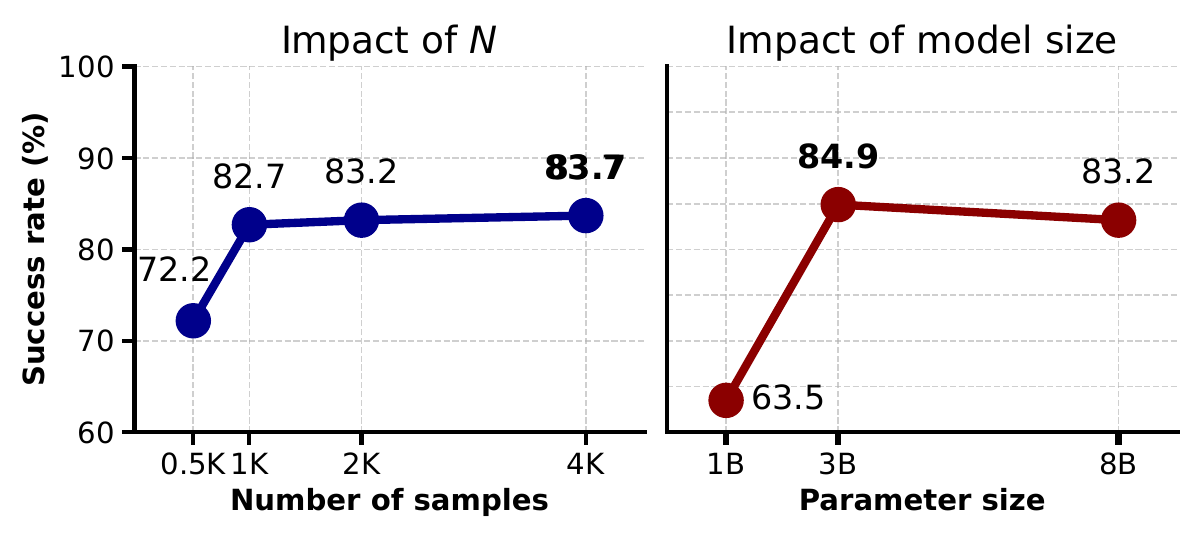}
\caption{\textbf{Ablation studies.} On the left, we study the impact of the number of samples $N$. We notice diminishing returns for our trained model and stop at 4K generated samples. On the right, we study the impact of the parameters of the model, proving that also smaller models can benefit from \framework.}
\label{fig:ablations}
\end{figure}

\subsection{Ablation studies}


In this section, we propose ablation studies. First, we study the impact of changing the number of per-task training samples $N$ that we use for training $\text{LLM}_\text{\framework}$. Second, we analyze if smaller models are compatible with \framework.
\medskip

\noindent\textbf{Training dataset size.}
In \framework, we can generate arbitrarily large datasets of training samples. To understand the impact of the data scale, we train LLaMA-8B with \framework on 4 different datasets generated automatically, containing 500, 1000, 2000, and 4000 samples per task ($N$). We report in Figure~\ref{fig:ablations} (left) the average results across the 9 tasks in simulation that we use in Table~\ref{tab:task_comparison}. From our results, it is evident that the size of $N$ significantly impacts the efficacy of the trained $\text{LLM}_\text{\framework}$, with a substantial loss of performance when $N=500$ (success rate \textbf{72.2\%}). However, we also noticed diminishing returns at the increase of $N$. In particular, we noticed that the performance doubling $N$ from 2000 to 4000 is only \textbf{+0.5\%}, so we have chosen $N=2000$ in the paper to shorten training times. In conclusion, we advocate that although \framework is effective in exploiting the synthetically generated data, more strategies may be needed to increase accuracy beyond saturation.
\smallskip

\noindent\textbf{Model size.}
Within \framework, we finetune a small model (LLaMA-8B) and use a larger one for data generation (LLaMA-70B). We aim to understand whether even smaller models can be trained with \framework, to enable applications on edge devices with low computational resources.
To do so, we trained LLaMA-3.2 1B and LLaMA-3.2 3B~\cite{meta2024llama} with \framework and compared it with our finetuned LLaMA-8B used for all other experiments. All models use data generated by LLaMA-70B as $\text{LLM}_\text{boot}$. We present results in Figure~\ref{fig:ablations} (right). The usage of the 3B and 1B models still results in remarkable performance. Interestingly, LLaMA-3.2 3B achieves \textbf{84.9\%} as average success rate, even marginally higher than LLaMA-8B (\textbf{83.2\%}). Please note that LLaMA-3.2 is a different release from LLaMA-8B (3.1), therefore, we attribute the higher performance to the superior data quality used for the 3.2 LLaMA release~\cite{meta2024llama}. This shows that even a  compact model used as $\text{LLM}_\text{BLAZER}$ can result in competitive performance with state-of-the-art zero-shot manipulation methods based on LLMs. Conversely, the 1B model yields lower successes (63.5\%), but still beats some baselines in Table~\ref{tab:task_comparison} with a very limited number of parameters. This shows further flexibility of \framework in model size for applications with significant computational constraints.

%% file: sections/conclusions.tex
\section{Conclusions}

In this paper, we introduced \framework, a  method for finetuning standard LLMs to obtain specialized agents for robotic manipulation. We demonstrated the efficacy of \framework in both simulated and real environments and evaluated its generalization performance on different tasks beyond those used for training. We believe that our work will encourage further research on the usage of pretrained LLMs for robotics-oriented tasks. While \framework currently exploits only positive demonstrations for supervised finetuning, future work can benefit from both positive and negative samples to learn more accurate manipulation models.
Indeed, unsuccessful episodes generated by our method in simulation could be used for more advanced post-training techniques such as Direct Preference Optimization (DPO)~\cite{rafailov2023direct}, that are able to exploit preference pairs including negative examples.